\documentclass[11pt]{article}

\usepackage[final]{acl}

\usepackage{times}
\usepackage{latexsym}

\usepackage[T1]{fontenc}

\usepackage[utf8]{inputenc}

\usepackage{microtype}

\usepackage{inconsolata}

\usepackage{graphicx}

\usepackage{booktabs}
\usepackage{comment}
\usepackage{lipsum}
\usepackage{scalefnt}
\title{Vision-Language Models Align with\\Human Neural Representations in Concept Processing}

\author{Anna Bavaresco, Marianne de Heer Kloots,\\\textbf{Sandro Pezzelle, Raquel Fern\'andez}\\Institute for Logic, Language and Computation \\ University of Amsterdam \\ \texttt{ \{a.bavaresco, m.l.s.deheerkloots, s.pezzelle, raquel.fernandez\}@uva.nl}}

\begin{document}
\maketitle
\begin{abstract}
Recent studies suggest that transformer-based vision-language models (VLMs) capture the multimodality of concept processing in the human brain. However, a systematic evaluation exploring different types of VLM architectures and the role played by visual and textual context is still lacking. Here, we analyse multiple VLMs employing different strategies to integrate visual and textual modalities, along with language-only counterparts. We measure the alignment between concept representations by models and existing (fMRI) brain responses to concept words presented in two experimental conditions, where either visual (pictures) or textual (sentences) context is provided. Our results reveal that VLMs outperform the language-only counterparts in both experimental conditions. However, controlled ablation studies show that only for some VLMs, such as LXMERT and IDEFICS2, brain alignment stems from genuinely learning more human-like concepts during \textit{pretraining}, while others are highly sensitive to the context provided at \textit{inference}. Additionally, we find that vision-language encoders are more brain-aligned than more recent, generative VLMs. Altogether, our study shows that VLMs align with human neural representations in concept processing, while highlighting differences among architectures. We open-source code and materials to reproduce our experiments at: \url{https://github.com/dmg-illc/vl-concept-processing}.


\end{abstract}

\section{Introduction}
\label{sec:intro}
Recent attempts to augment language-only models with vision have resulted in a multitude of vision-language models (VLMs), integrating modalities with different strategies based on the task at hand. While the practical implications of multimodality are evident---it allows language-only models to `see' and perform novel vision-language tasks---, its theoretical implications from the point of view of human language modelling are not yet fully understood. 

Insights from cognitive and neuroscientific work suggest that human semantic representations are deeply grounded in multimodal sensory experiences
\cite{louwerseSymbolInterdependencySymbolic2011,barsalou1999perceptual,harnad1990symbol,bergen2012louder}, and that visual and linguistic stimuli evoke shared neural activity patterns \citep{devereux2013representational, simanova_modality-independent_2014, popham2021visual, kaup2024modal}. These findings motivate the hypothesis that semantic representations learnt by VLMs approximate the human ones better than representations by unimodal (both language-only and vision-only) models.

However, existing works testing this empirically point to diverging conclusions, with some studies documenting advantages of multimodality \cite[e.g.,][]{oota-etal-2022-neural, zhuang-etal-2024-lexicon}, others suggesting that multimodal fine-tuning `harms' the human-alignment properties originally acquired by language models \cite[e.g.,][]{bavaresco-fernandez-2025-experiential}, and more nuanced contributions indicating that the advantages of multimodality are circumscribed \cite[e.g.,][]{pezzelle2021word, dong2023vision, zhuang-etal-2024-visual, ryskina2025language}.
A likely reason behind the difficulties in reconciling the findings from these studies lies in their addressing similar research questions with different methods.

Among others, two under-investigated experimental factors that may strongly impact measures of human-likeness computed on VLM semantic representations concern the type and amount of context provided, and the differences between VLM architectures. More concretely, many psycholinguistic studies on multimodality in concept processing analysed human measures (both behavioural and neural) collected by presenting participants with non-contextualised words \cite[e.g.,][]{pezzelle2021word, zhuang-etal-2024-visual, bavaresco-fernandez-2025-experiential}. This may have resulted in limited engagement of multimodal knowledge, ultimately leading to an under-appreciation of VLMs' brain modelling potential. An additional limitation of previous work is that it mostly focused on the multimodal/unimodal dichotomy, placing little emphasis on the substantial differences existing between multimodal architectures, and their implications for brain modelling. 

In this study, we systematically compare ten models, including VLMs and language-only counterparts, by measuring their correlation with a subset of the Pereira dataset \cite{pereira2018toward}, collecting human fMRI (functional magnetic resonance imaging) responses to 180 concepts. Our experiments address the limitations of previous work in two respects.

\begin{figure*}
    \centering
    \includegraphics[width=0.95\textwidth]{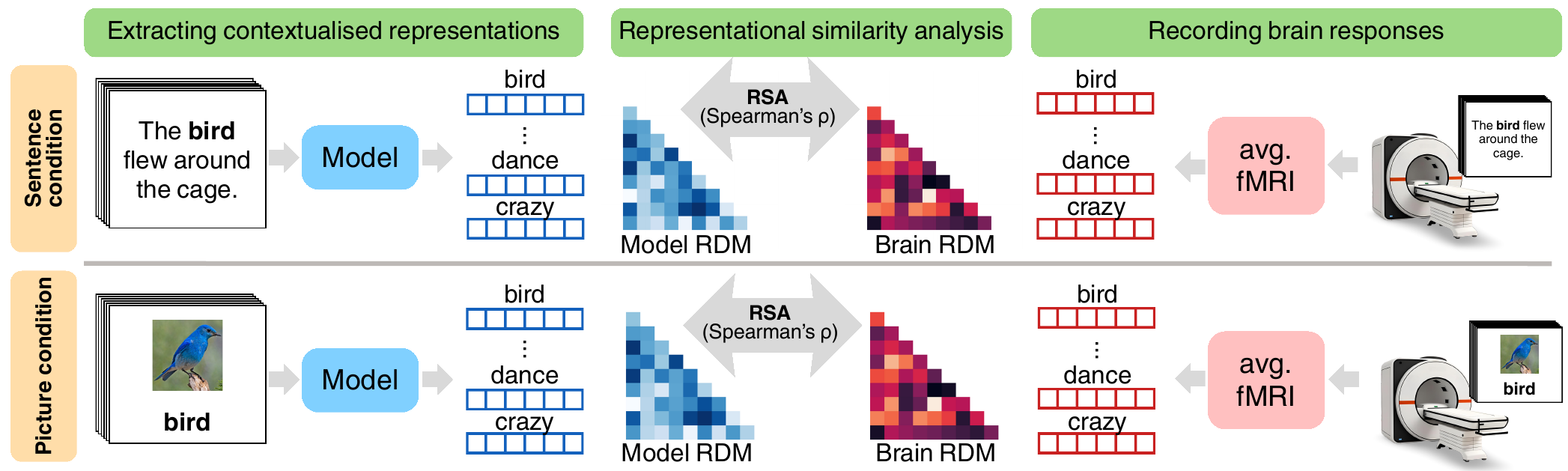}
    \caption{Overview of the experimental setup in the sentence (top) and picture (bottom) condition. Models are fed with the same stimuli participants saw in the fMRI scanner, i.e., concept words appearing in six contexts (provided by either sentences or pictures). Note that contexts are intended to highlight the same word meaning, but may describe different situations (sentences are \textit{not} image captions). Model representations and brain responses averaged across the six contexts are then used to derive representational dissimilarity matrices (RDMs), storing pairwise cosine distances. Finally, the Spearman correlation between these RDMs provides a measure for model--brain alignment. Best viewed in colour.
    }
    \label{fig:experiments-diagram}
\end{figure*}

First, we explore the role played by the \textbf{context} provided through different modalities; we analyse brain responses to concept words that were presented to participants with either a sentential context (\textit{sentence condition}, where concept words are presented within sentences) or a visual one (\textit{picture condition}, where concept words are accompanied by images illustrating their content), as shown in Figure~\ref{fig:experiments-diagram}. Besides ensuring that multimodal knowledge is engaged, focusing on this dual-context setup allows us to better characterise the brain-aligning properties of VLMs: if VLMs are more brain-aligned than their language-only counterparts only when visual context is provided, their superiority may be simply due to accessing visual information that language-only models cannot `see'; on the other hand, if VLMs have an advantage even without receiving an image input, this signals they have truly learnt human-like, multimodal concept representations during pretraining.

Second, we evaluate different \textbf{VLM families}: we indentify three classes of VLMs integrating the visual and textual modalities through different strategies, and test two representative exemplars for each. By comparing their alignment with brain responses, we test whether multimodality leads to a generalised pattern of improvement or is family/architecture-dependent.

Our results reveal that VLMs tend to be more brain-aligned than their language-only counterparts in both context conditions, suggesting that they successfully model human-like concept processing. In addition, we find that vision-language encoders, such as VisualBERT \cite{li2019visualbert} and LXMERT \cite{tan2019lxmert}, model brain responses better than the other VLM types, including the more powerful, generative ones. This means that performance on downstream tasks, where generative VLMs excel, may not go hand in hand with human-like concept processing. More broadly, our findings contribute to the ongoing debate around whether multimodality results in more human-like language models or not, showing that it is beneficial for brain alignment in concept processing. 

\section{Related Work}
\label{sec:rel_work}
\subsection{Multimodal models in cognitive modelling}
\label{sec:mm_in_cm}

Several recent studies have used transformer-based multimodal models in brain encoding/decoding experiments or to model behavioural data.
Here, we review the closest to our focus, i.e., those analysing responses to concepts as opposed to more complex stimuli (e.g., videos).

\citet{pezzelle2021word} evaluated word representations extracted from several VLM architectures against human semantic judgments, including word similarity and relatedness benchmarks. Their findings reveal that VLM word representations align better with human judgments than representations from text-only models, although only on concrete word pairs. Similarly, \citet{zhuang-etal-2024-lexicon} and \citet{zhuang-etal-2024-visual} found visual grounding to result in improved and more efficient word learning, especially in low-data regimes. 

Considering studies analysing brain activations, \citet{oota-etal-2022-neural} used several unimodal (language-only and vision-only) models and VLMs to predict a subset of the Pereira dataset (only the picture condition), and found a VLM \cite[VisualBERT,][]{li2019visualbert} to be more brain-predictive than the other architectures.
Crucially, in this setup, multimodal representations are the only ones receiving the same input shown to the human participants (i.e., image+word). Therefore, it is difficult to determine if their advantage is due to capturing something fundamental about concept processing, or simply to having access to more information than unimodal models.

\citet{bavaresco-fernandez-2025-experiential} partially addressed this limitation by comparing multimodal and language-only models in a controlled setup, where they were all fed with text-only inputs and used to model fMRI responses collected while participants viewed isolated nouns. Their results, surprisingly, reveal an advantage of language-only models, even on a subset of more concrete nouns. As it is known that the presence of linguistic context influences both brain responses to concepts \cite{xu2005language, deniz2023semantic} and mental simulations of their content \cite{zwaan2014embodiment}, a likely explanation for their findings is that the brain responses they analysed did not reflect a detectable engagement of multimodal knowledge. 

Finally, \citet{ryskina2025language} analysed, again, the Pereira dataset and predicted fMRI responses using both VLMs and large language-only models (LLMs). Their main goal was to identify concept-sensitive brain regions and check whether their neural activations can be successfully modelled with VLMs and LLMs. Systematic comparisons of VLM families and analyses of the role played by context were, therefore, beyond the scope of their study.\looseness-1

We complement these works by analysing brain responses to concepts presented within different context conditions, including a picture condition, as analysed by \citet{oota-etal-2022-neural}, and a sentence condition. This allows us to determine if VLMs' alignment with human brain responses is modulated by the context or consistent across visual and sentential contexts.

\subsection{VLM families}
Transformer-based vision-language models (VLMs) can be divided into three categories: contrastive models, vision-language encoders, and generative VLMs. Contrastive models, such as CLIP \cite{radford2021learning}, ALIGN \cite{jia2021scaling}, and LiT \cite{Zhai_2022_CVPR}, encode images and text separately, with two dedicated transformer-based modules. These modules are pre-trained with a contrastive loss, which maximises the similarity between image and text embeddings of matching image-text pairs.

Vision-language encoders are characterised by a specific module that uses attention mechanisms to learn relations between visual features extracted with an object detector and text embeddings. In some architectures, such as VisualBERT \cite{li2019visualbert}, language processing and multimodal integration are performed by the same BERT-based \cite{devlin-etal-2019-bert} module; in other architectures, e.g., LXMERT \cite{tan2019lxmert}, text is encoded in a dedicated transformer module before being passed to the cross-modal module.

Lastly, generative VLMs consist of a pretrained LLM, a pretrained vision encoder, and an `adaptor' (or `projector'), i.e., a shallow module which learns a mapping between the space of the image tokens and that of the language tokens. Examples of these architectures are LLaVA-NeXT \cite{liu2024llavanext}, IDEFICS2 \cite{laurenccon2024matters}, Qwen2.5-VL \cite{bai2025qwen2}, and Molmo \cite{deitke2024molmo}.\footnote{Here, we do not review proprietary models as they do not allow extracting layer-wise representations, which are fundamental for assessing brain alignment.}

The studies reviewed above (Section~\ref{sec:mm_in_cm}) used exemplars from at most two of the VLM families, but a comprehensive comparison remains elusive. To address this gap, we experiment with six different VLMs, including representatives from each family.\looseness-1


\section{Methods}
\label{sec:methods}

To study the alignment between vision-language models (VLMs) and brain responses, we focus on a dataset of neural responses to concept words, which participants read either accompanied by pictures or within sentences (Experiment 1 in \citealp{pereira2018toward}). We derive representations from the same concept words using a set of VLMs and language-only models and quantify their alignment with human responses in two brain networks with representational similarity analysis \cite[RSA,][]{kriegeskorte2008representational}. 
An overview of our experimental setup is provided in Figure~\ref{fig:experiments-diagram}.

\subsection{Brain responses}
The brain responses we focus on were collected by \citet{pereira2018toward}. They consist of
voxel-wise fMRI activations collected while 16 participants were presented with English words 
representing specific concepts in different conditions. There are 180 concept words in total, 
including different parts of speech (128 nouns, 22 verbs, 29 adjectives and adverbs, 
and 1 function word; see Appendix~\ref{sec:appendix_pereira_dataset} for the full list). 

We consider two experimental conditions: a language-only \textit{sentence condition} and a 
multimodal \textit{picture condition}. In the sentence condition, participants were fMRI-scanned 
while reading sentences where the target words were boldfaced. For each concept word, participants saw 
six sentences, one at a time. In the picture condition, each word was presented together with an 
image illustrating the relevant concept. Again, participants 
viewed six different images for each concept word, one at a time. In both conditions, participants were asked to think about the meaning of the target concept. 

We use the brain responses as preprocessed by \citet{pereira2018toward}. More concretely, the preprocessed response for each stimulus consists of an array where entries represent the `magnitude' of the brain activity at different voxels. While responses were recorded for the whole brain, we focus on two specific regions, 
involved in either linguistic or visual processing: 
the left-hemisphere \textit{Language network} \citep{fedorenko2011functional} and 
the \textit{Visual network} \citep{power2011functional, buckner2008brain}. Responses are averaged across the six presentations of the same word per condition; hence, for each participant, we have one averaged brain response per concept, condition, and brain network. See Appendix~\ref{sec:brain_data} for additional details on the brain responses and the anatomical regions included in the two brain networks.


\subsection{Models}

We employ two main types of models:  
a set of VLMs trained on images and text, and 
a set of language-only models trained exclusively on text.
To make our evaluation comprehensive, 
we include multiple VLMs that are representative of the families described in Section~\ref{sec:rel_work}. 
Regarding the language-only models, 
we include architectures 
that either provide informative baselines (e.g., GloVe) 
or useful comparisons with specific VLMs.\footnote{Note that testing state-of-the-art models is not crucial for our goals. We choose fully open-source VLMs that have a language-only counterpart and were among the best ones when the experiments were conducted.} We briefly describe the models here and refer to Appendix~\ref{sec:model_details} for details about the specific implementations. 

\paragraph{Vision-language models} We select two representative VLMs from each family. More specifically, we
consider CLIP \citep{radford2021learning} and ALIGN \citep{jia2021scaling} for the
contrastive VLMs, VisualBERT \citep{li2019visualbert} and LXMERT \citep{tan2019lxmert}
for the vision-language encoders, and IDEFICS2 \citep{laurenccon2024matters} and 
LLaVA NeXT \cite{liu2024llavanext} for the generative VLMs. 

\paragraph{Language-only models}
 
We experiment with one encoder-only language model \cite[BERT,][]{devlin-etal-2019-bert} and 
two decoder-only large language models---Mistral \cite{jiang2023mistral} and Llama3 \cite{llama3}. 
In addition, we include the simpler distributional semantic model GloVe \citep{pennington2014glove}. 
BERT
forms the basis of the language components of ALIGN, VisualBERT and LXMERT,\footnote{While VisualBERT 
is initialised with BERT weights, LXMERT and ALIGN 
train their BERT-based modules from scratch.} while Mistral and Llama3  are the language models used in IDEFICS2 and LLaVA NeXT, respectively. Finally, while GloVe does not output contextualised representations, its embeddings were used by \citet{pereira2018toward} to select the concept stimuli for the fMRI study; 
we therefore include them as they provide a useful reference.

\subsection{Extracting representations}
\label{sec:procedure}

To extract model representations that we can compare to the fMRI responses, 
we feed the models with the same stimuli presented to the participants in 
the experiments by \citet{pereira2018toward}. Different models require 
slightly different inputs, as we explain in detail in this section. 

\paragraph{Sentence condition}

In this condition, participants read sentences without visual information. All transformer-based models (both VLMs and language-only models) are fed with the same sentences, without any image. 
One exception to the procedure is LXMERT; as it requires some visual input in addition to text, we pass  
an image made up of random noise along with each sentence.\footnote{We create a different random-noise image 
for each sentence. These images are linked to our public GitHub repository.} 

After extracting model representations (\textit{hidden states}) for the entire sentences from each model layer, we select those corresponding to the tokens of the target concept word. If the target word consists of several tokens, we average the hidden states of the relevant tokens. Finally, we average again over the six contexts where the word appears. All averages are computed layer-wise. This means that our procedure yields one concept representation for each concept at each model layer.


As for GLoVe, given that it provides static (type-based) 
decontextualised representations, we simply select the pre-trained 
GloVe embedding corresponding to each target word.

\paragraph{Picture condition}

In this condition, participants see each word together with an image. 
For the VLMs, both the target word and each image accompanying it are fed 
to the models. When using contrastive models, we integrate the image and text 
embeddings by taking their element-wise sum.\footnote{We also tested element-wise multiplication and concatenation, but these resulted in less brain-aligned representations.} For the VL encoders, we extract 
hidden states from all text-specific layers and from the cross-modal layers. 
In generative VLMs, we consider the hidden states from all layers of the LLM module.
To obtain a final representation for each concept word that can be compared 
to brain activations, we then average the extracted representations across the six images per word. 
For the language-only models, only the target word is provided as input. 

\begin{figure*}
\centering
\begin{tabular}{c}
     \includegraphics[width=0.85\textwidth]{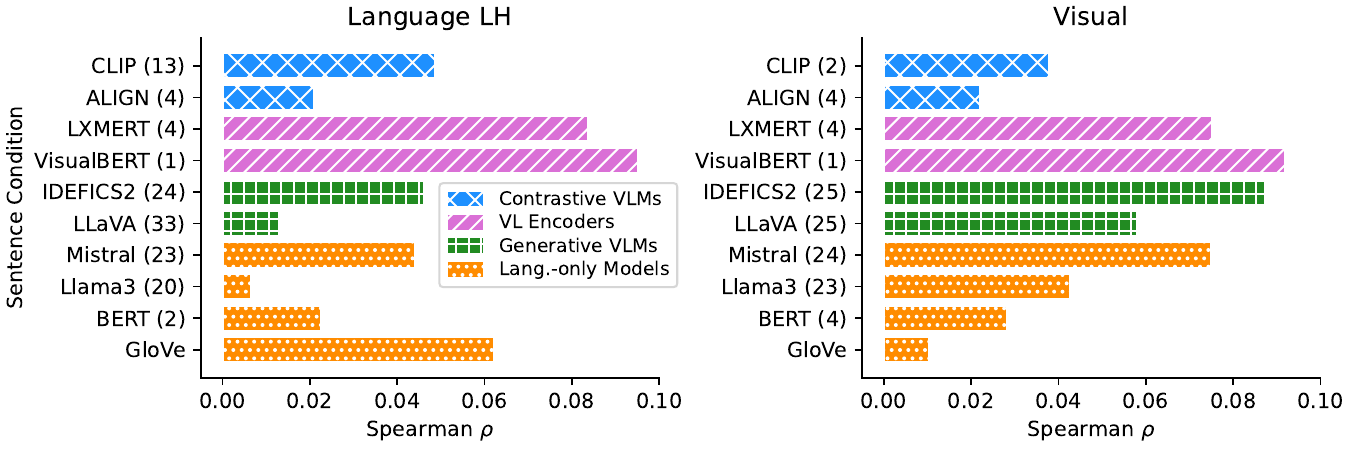}   \\
     \includegraphics[width=0.85\textwidth]{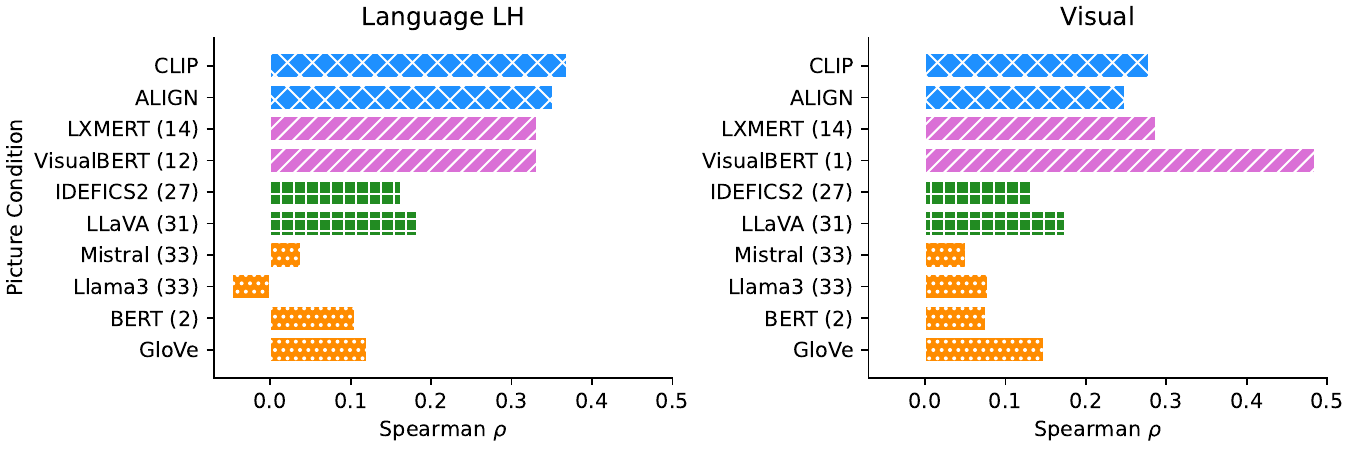} 
\end{tabular}

    \caption{RSA results for the \textit{sentence condition} (upper row) and \textit{picture condition} (lower row). Spearman correlations indicate the alignment between concept representations by models and fMRI responses in the left-hemisphere (LH) language network and in the visual network. Numbers in brackets indicate the model layer from which representations were extracted. Note that the range of the $x$ axes differs between conditions.}
    \label{fig:brain-rsa}
\end{figure*}

\subsection{Evaluation}
We evaluate the models' alignment to fMRI data with Representational 
Similarity Analysis \citep[RSA,][]{kriegeskorte2008representational}, which is illustrated in Figure~\ref{fig:experiments-diagram}. In the following, we explain how it differs from other existing metrics and describe how it was computed in our setup.

\paragraph{RSA vs.~other metrics} Given two representational spaces (e.g., those defined by fMRI responses and model representations), RSA measures whether the between-stimuli relations (typically distances) in one space are similar to the relations in the other \cite[\textit{second-order isomorphism},][]{shepard1970second}. This means that the information it provides is at the level of similarity between representational geometries (\textit{representational similarity}), rather than between feature spaces \cite{dujmovic2024inferring}.

Alternative metrics that are better suited to study brain predictivity at the feature level concern the `brain encoding' performance, or `linear predictivity' \cite{naselaris2011encoding}. These methods involve training transformations from the model embedding space to optimise the fit with neural data, which usually results in higher correlation scores than RSA. However, such metrics have been found lacking in other desirable properties such as functional consistency \cite[e.g.,][]{boEvaluatingRepresentationalSimilarity2025}. Additionally, the fact that they involve an optimisation step makes it difficult to disentangle how much the resulting correlations are attributable to model features vs.~optimisation.

Given that the best choice of alignment metric ultimately depends on the research goals \cite{ivanovaProbingArtificialNeural2021a}, we deem RSA to be suitable for our interest in observing relative differences between models rather than maximising neural predictivity.

\paragraph{Computing RSA} As routinely done in RSA, we approximate brain and model spaces through representational dissimilarity matrices (RDMs), whose entries store cosine distances between concept representations (i.e., brain responses or model embeddings). We then quantify their alignment as a Spearman correlation over the vectorised off-diagonal elements of the RDMs, corresponding to the unique set of pairwise distances in each space. 
A significant positive correlation between RDMs indicates that there is representational similarity between brain-derived and model-derived spaces. We compute representational similarity against brain RDMs 
averaged across single participants.\footnote{We elaborate more on the reasons behind this choice in the Limitations section at the end of the paper.} 
For each model with multiple layers, 
RSA is computed separately for each layer's representations.
When drawing comparisons between models, we consider the best (i.e., most aligned) 
layer for each model. We verify that differences in brain alignment between models 
are statistically significant by applying a Fisher transformation 
to all the unique pairs of Spearman correlations and calculating 
the $p$-value associated with the difference between the two $z$-scores. 
To control for false positives due to multiple comparisons 
(45 per brain network in each condition), 
$p$-values are Bonferroni-corrected with $\alpha$ = 0.05.
 We report the complete set of $p$-values in Appendix~\ref{sec:model_comparisons}, Table~\ref{tab:p_vals}. To obtain a random baseline for the representational similarity, 
we compute alignment against shuffled brain responses, i.e., 
a condition where each concept was associated with a randomly chosen 
fMRI response from a non-matching concept.

\section{Results}
\label{sec:results}

Our main results are 
visualised in Figure~\ref{fig:brain-rsa},
which reports representational similarity from the most brain-aligned model 
layer in the cases where we extract representations from multiple. All the displayed correlations are significantly different from 0, which coincides with the random baseline computed by shuffling brain responses.
Layer-wise RSA values for all models are provided in Appendix~\ref{sec:appendix_layerwise_results}. 

\paragraph{Sentence condition}

As displayed in the upper row of Figure~\ref{fig:brain-rsa},
all models exhibit low to moderate positive correlations with brain responses from both networks. 
In the language network, the VL encoder VisualBERT is 
statistically significantly more brain-aligned than all the other 
models ($\rho$ = 0.10) except for LXMERT ($\rho$ = 0.08). 
Additionally, GloVe's correlation with fMRI responses is not significantly different from that of the more advanced models---both multimodal and unimodal---CLIP, LXMERT, IDEFICS2 and Mistral. 
Comparisons between VLMs and their language-only counterparts reveal that both VisualBERT and LXMERT are significantly more brain-aligned than BERT, while LLaVA and IDEFICS2 are not different from their language decoders Llama3 and Mistral.

In the visual network, language-only models surprisingly achieve significant correlations, with Mistral performing comparably with the VLM LXMERT and outperforming LLaVA, CLIP and ALIGN. However, the highest correlations are still observed for the VLMs VisualBERT and IDEFICS2 ($\rho$ = 0.09 for both), which significantly outperform all other models except for LXMERT and Mistral. Regarding the remaining comparisons between VLMs and language-only counterparts, LLaVA's advantage over Llama3 is not significant, while both VisualBERT and LXMERT are significantly more brain-aligned than BERT. Remarkably, language-only models achieve significant correlations in this network even if no visual information is being presented. 


\paragraph{Picture condition}

Results displayed 
in the lower row of Figure~\ref{fig:brain-rsa} show higher correlations 
than in the sentence condition, which can be attributed to higher signal 
(i.e., higher inter-participant similarities) in the fMRI responses. Notably, VLMs are significantly more brain-aligned than their unimodal counterparts in both brain networks.

In the LH language network, both contrastive VLMs and VL encoders exhibit moderate correlations
with brain responses ($\rho$ > 0.3), which are all statistically significantly stronger 
than those achieved by generative VLMs and language-only models.  
In the visual network, VisualBERT outperforms all other models ($\rho$ = 0.48) 
and CLIP, ALIGN and LXMERT exhibit similar brain correlations (0.25 < $\rho$ < 0.29). 

Lastly, GloVe exhibits higher correlations than all other more powerful language-only models in the language network. This is likely because \citet{pereira2018toward} selected the concept stimuli to include in their experiment using GloVe representations, as we mentioned above.

\subsection{Trends across the board}

Overall, our results indicate an advantage of VLMs over the language-only counterparts that is consistent across conditions and brain networks. Below, we highlight the main implications regarding differences among VLMs and comparisons with their unimodal counterparts. 

\paragraph{VLM families} Our findings reveal that VL encoders are the most brain-aligned, outperforming the other model families in all scenarios except for the language network in the picture condition, where contrastive VLMs have an advantage. A possible explanation for this result is that contrastive VLMs are effective at capturing object-word 
correspondences---a scenario quite akin to the picture condition---but struggle to accurately represent more complex relations between entities \cite{winoground, vsr, valse, svo_probing}. 
This limitation may, therefore, have resulted in lower performance compared to the VL encoders.

Regarding generative VLMs, their brain correlations in the picture condition are \textit{lower} than those by contrastive VLMs and VL encoders. Even in the sentence condition, where there is linguistic context they can incorporate in their word representations, they still do not outperform VL encoders---a surprising result, given their superior performance on downstream tasks. A potential reason for this 
may be that autoregressive pretraining privileges production over representation \cite{behnamghader2024llm2vec, muennighoff2025generative, Springer2024RepetitionIL}, making \textit{word}-level representations less expressive than those by previous architectures.




\paragraph{VLMs vs.~language-only counterparts} In general, we observe an advantage of VLMs over language-only models, albeit with differences between conditions.
In the picture condition, all VLMs are more brain-aligned than their unimodal counterparts, indicating that their ability to process visual information is beneficial for modelling brain-relevant semantic aspects. 

In the sentence condition, results are more nuanced, with VisualBERT and LXMERT outperforming BERT, and IDEFICS2, LLaVA and ALIGN never significantly outperforming Mistral, Llama3 and BERT, respectively. This suggests that the type of context provided (visual vs.~sentential) affects the brain-modelling abilities of VLMs~vs. their language-only counterparts. We further investigate this aspect in two ablation studies. 

\section{Ablation Studies}
\label{sec:additional_experiments}

To complement the findings provided by RSA, we conduct 
two ablation analyses aimed at better understanding what drives VLMs' 
brain alignment in both experimental conditions. 

\subsection{Semantic information in the sentence condition}
The analyses presented earlier allow comparing the brain alignment of VLMs against that of the language-only counterparts. However, finding that VLMs achieve similar brain alignment to unimodal counterparts does not automatically indicate they rely on the same knowledge: while it seems natural to assume that representations by a VLM will incorporate the brain-relevant information by the underlying language module (and perhaps learn additional information as a result of multimodal fine-tuning), it is still possible that vision-language fine-tuning alters the language-module information substantially. The two scenarios are illustrated in Appendix~\ref{sec:partial_correlations}, Figure~\ref{fig:venn-diagram}.
To investigate this, we conduct a partial correlation analysis aimed at removing from VLMs' embedding spaces the information shared with LLMs' spaces. We focus on a subset of models whose architecture is highly similar to that of the language-only counterpart, i.e., VisualBERT, LXMERT, LLaVA and IDEFICS2.\footnote{While VisualBERT, LLaVA and IDEFICS2 were initialised with pretrained weights from their language-only counterparts,  ALIGN and LXMERT were trained multimodally from scratch and are hence not included.} 

\paragraph{Computing partial correlations} Partial correlations are computed as follows. 
Consider the RDM from a VLM $y$, the RDM from its language-only counterpart $x$, 
and the residuals $r_{i} = y_{i} - \hat{y}_{i}$ from the linear 
regression with equation $\hat{y}_{i} = a + bx_{i}$. 
The partial correlation achieved by the VLM is defined as 
the Spearman correlation $\rho(r_{i},z_{i})$, where $z$ 
represents the RDM of the brain responses, and it provides an indication of the VLM's brain alignment achieved once the representational information shared with the language-only counterpart is ablated.\looseness-1 

\begin{figure}[!t]
    \centering
    \includegraphics[width=0.85\linewidth]{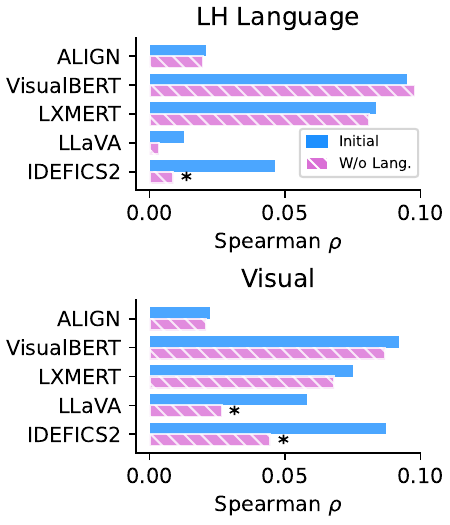}
    \caption{\textcolor{cyan}{Initial} (as reported in the main experiment) and \textcolor{magenta}{partial} correlations between VLM representations and fMRI responses in the \textit{sentence condition}. Statistically significant differences (marked by asterisks) between initial and partial correlations indicate that the brain-relevant information captured by the VLM is shared with that present in its language module.}
    \label{fig:partial_correlations_sentence}
\end{figure}

\paragraph{Results}
VLMs' initial (as reported in the main experiment) and partial correlations with brain responses are displayed in Figure~\ref{fig:partial_correlations_sentence}. If the brain alignment of a VLM is driven by its language-only component, the partial correlation should be significantly lower than the initial correlation, signalling that the removed information mattered. In contrast, the absence of such a difference indicates that the semantic knowledge learnt by the language-only module was not responsible for the brain alignment achieved by the VLM; hence, the knowledge driving alignment must come from multimodal training.\looseness-1 

In the language network, all differences between partial and initial correlations---except for IDEFICS2 in language LH---are \textit{not} 
statistically significant, suggesting that the brain alignment achieved by these models 
cannot be attributed to semantic information already present in the language-only modules, 
but to different knowledge acquired during multimodal pretraining.  
In the visual network, results reveal diverging patterns between generative VLMs and the other VLMs: the differences between initial and partial correlations are statistically significant for the former, while they are not for the latter. 
Interestingly, this suggests that part of the information relevant for alignment with visual brain responses was already present in Mistral and Llama3 before any vision-language training. That is, the brain-aligning information in these language-only models (see Figure \ref{fig:brain-rsa}) seems to be maintained after multimodal fine-tuning.

\begin{figure*}[!t]
    \centering
    \includegraphics[width=0.9\linewidth]{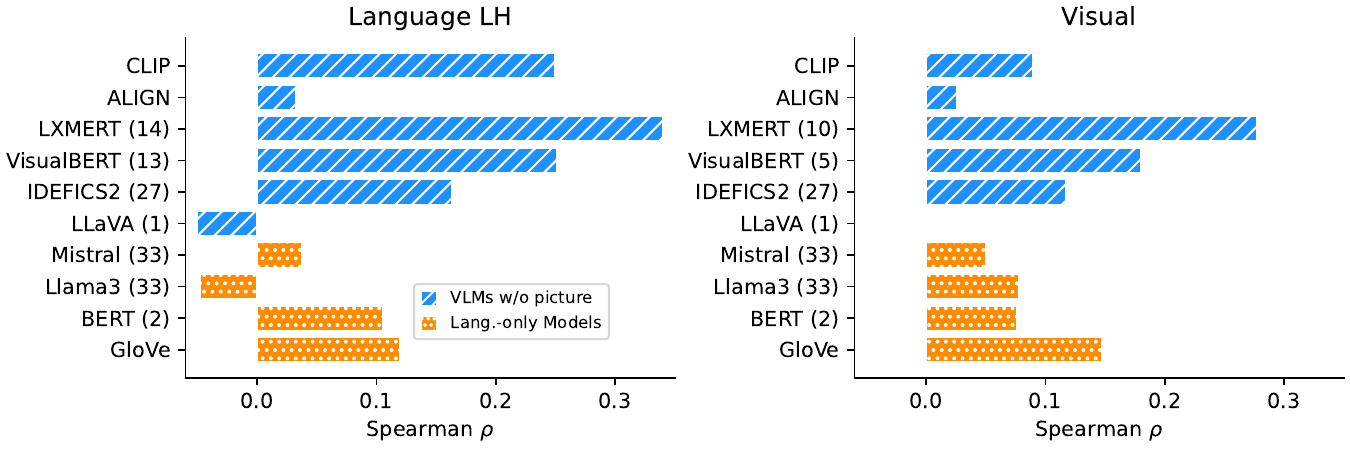}
    \caption{Results from the ablation study where we pass only concept words to both VLMs and language-only models. For both brain networks, we show the Spearman correlations resulting from RSA, indicating the alignment between models and fMRI responses from the \textit{picture condition}. Numbers in brackets indicate the layers from which representations are extracted.}
    \label{fig:abl_picture}
\end{figure*}
\subsection{Visual information in the picture condition}
\label{sec:ablation_study}
In the picture condition, we found a systematic advantage of VLMs over their language-only counterparts. However, a potential confound could be that VLMs have access to additional contextual information---the picture---that is available to human participants but not to language-only models. In this sense, their superior alignment with human responses could stem from an uneven comparison rather than indicate that they capture additional semantic information. 

To shed more light on this issue, we conduct an ablation study where we pass only the concept word (without pictures\footnote{Again, LXMERT requires a visual input. As we do in the sentence condition, we pass a random-noise image.}) to the VLMs, so that the input they receive matches that provided to language-only models. 
We then recompute RSA against human fMRI responses from the picture condition, following the same procedures employed in the main experiment. The resulting Spearman correlations are provided in Figure~\ref{fig:abl_picture}. We assess the statistical significance of differences between each pair of correlations per network as in the main experiment. 

This analysis reveals striking differences between VLMs: CLIP, ALIGN, VisualBERT, and LLaVA suffer dramatic drops in brain alignment when the picture is not provided as input, indicating their initial correlation was highly influenced by the context provided by the picture; on the other hand, brain-alignment variations observed for LXMERT and IDEFICS2 are minimal, suggesting their brain alignment is less sensitive to the presence of an input image (see also Figure~\ref{fig:pic-cond-ablation} in the Appendix).


Despite the performance drop suffered by some of the VLMs, the most brain-aligned architectures in both networks remain multimodal: LXMERT is significantly more brain-aligned than all other models, and VisualBERT significantly outperforms all language-only models.

Considering VLMs and their language-only counterparts, LXMERT and VisualBERT significantly outperform BERT, while ALIGN is \textit{less} brain-aligned than BERT across both brain networks. As for the generative VLMs, IDEFICS2 remains significantly more brain-aligned than Mistral, and LLaVA is significantly \textit{less} brain-aligned than LLama3 in the visual network.

Overall, this analysis suggests that, while some architectures rely heavily on input images, others yield strong brain correlations even without meaningful visual input.



\section{Conclusion}
\label{sec:conclusion}

We provide a broad investigation of the brain alignment to human concept processing achieved by vision-language models from different families, drawing meaningful comparisons with language-only models and considering two experimental conditions and two brain networks. 

Our results show evidence that the highest brain alignment is consistently achieved by one of the VLMs (and not a language-only model), although not the same architecture across all conditions and brain networks. Additionally, we find that vision-language encoders tend to exhibit higher brain alignment than the more recent generative VLMs. Lastly, our findings demonstrate that the superior brain alignment achieved by vision-language encoders cannot be solely attributed to receiving additional input at inference, but stems from learning novel multimodal semantic information during their pretraining.

More broadly, our findings speak to a wider research effort aimed at better understanding the effects of multimodality on language modelling. This includes works showing that vision-language training harms performance on natural-language-understanding \cite{iki-aizawa-2021-effect} and commonsense-reasoning benchmarks \cite{Madasu_2023_CVPR}, and others finding it results in improved deployment of taxonomic knowledge \cite{qin2025visionandlanguage} and prediction of words in context \cite{wang2023finding}, but without fundamentally altering linguistic representations. We believe that our work, together with concurrent NeuroAI studies such as those reviewed in the previous sections, contributes an additional meaningful perspective, and we hope it will inspire future efforts to evaluate and design more human-like VLMs.


\section*{Limitations}
\label{sec:limitations}

Our aim was to provide a comprehensive evaluation of VLMs off-the-shelf, as they are being used by the AI community. It is important to consider that these VLMs differ along many dimensions, including size, architecture, learning objective and amount of training data. While the differences in brain alignment we observed between them are interesting and meaningful, our setup does not allow attributing them to one specific factor.

Another potential limitation of our study pertains to the ROIs we considered. The visual network and the left-hemisphere language network we analysed provide useful insights into how the brain responds to visuo-linguistic stimuli, but two caveats remain. The first is that the involvement of the language network in semantic processing is still under investigation, with recent studies suggesting that its activation may not be required for semantic processing \cite{ivanova2021language}, and that additional regions outside this network are activated by concepts irrespective of the modality through which they are presented \cite{ryskina2025language}. Relatedly, the second caveat is that part of the multimodal integration arising from concept processing may happen in brain regions that we did not analyse, but were previously shown to be involved in multimodal processing \cite{fairhall2013brain, bonner2013heteromodal, handjaras2016concepts, nikolaus2024modality}.

Finally, we did not investigate inter-participant differences in brain alignment. We chose to average brain responses (more precisely, \textit{brain RDMs}) across participants due to the high amount of noise in the individual data, especially in the language network. Averaging responses allowed us to get a stronger signal when measuring correlations with model representations. At the same time, this procedure may have `evened out' some participant-specific patterns.

\section*{Acknowledgments}
We thank the members of the Dialogue Modelling Group (DMG) from the University of Amsterdam and Leonardo Bertolazzi for the helpful feedback provided at different stages of this project.

We express our gratitude to the anonymous reviewers who offered valuable insights on the current and previous submissions. Their advice has contributed to improving the present work significantly.

A.~B.~and R.~F.~are funded by the European Research Council (ERC) under the European Union’s Horizon 2020 research and innovation programme (grant agreement No.\ 819455). M.~d.~H.~K.~is funded by the Netherlands Organisation for Scientific Research (NWO), through a Gravitation Grant 024.001.006 to the Language in Interaction Consortium.

\bibliography{custom}

@inproceedings{ivanovaProbingArtificialNeural2021a,
	title = {Probing artificial neural networks: insights from neuroscience},
	doi = {10.48550/arXiv.2104.08197},
        publisher = {ICLR 2021 Workshop: How Can Findings About The Brain Improve AI Systems?},
	author = {Ivanova, Anna A. and Hewitt, John and Zaslavsky, Noga},
	year = {2021}
}

@inproceedings{boEvaluatingRepresentationalSimilarity2025,
	title = {Evaluating {Representational} {Similarity} {Measures} from the {Lens} of {Functional} {Correspondence}},
	doi = {10.48550/arXiv.2411.14633},
	publisher = {Proceedings of the {C}ognitive {C}omputational {N}euroscience conference (CCN)},
	author = {Bo, Yiqing and Soni, Ansh and Srivastava, Sudhanshu and Khosla, Meenakshi},
	year = {2025},
}

@inproceedings{zhuang-etal-2024-lexicon,
    title = "Lexicon-Level Contrastive Visual-Grounding Improves Language Modeling",
    author = "Zhuang, Chengxu  and
      Fedorenko, Evelina  and
      Andreas, Jacob",
    editor = "Ku, Lun-Wei  and
      Martins, Andre  and
      Srikumar, Vivek",
    booktitle = "Findings of the Association for Computational Linguistics: ACL 2024",
    month = aug,
    year = "2024",
    address = "Bangkok, Thailand",
    publisher = "Association for Computational Linguistics",
    url = "https://aclanthology.org/2024.findings-acl.15/",
    doi = "10.18653/v1/2024.findings-acl.15",
    pages = "231--247"
}

@article{louwerseSymbolInterdependencySymbolic2011,
	title        = {Symbol {Interdependency} in {Symbolic} and {Embodied} {Cognition}},
	author       = {Louwerse, Max M.},
	year         = 2011,
	journal      = {Topics in Cognitive Science},
	volume       = 3,
	number       = 2,
	pages        = {273--302},
	doi          = {10.1111/j.1756-8765.2010.01106.x},
	issn         = {1756-8765},
	url          = {https://onlinelibrary.wiley.com/doi/abs/10.1111/j.1756-8765.2010.01106.x},
	urldate      = {2024-06-25},
	copyright    = {Copyright © 2010 Cognitive Science Society, Inc.},
	note         = {\_eprint: https://onlinelibrary.wiley.com/doi/pdf/10.1111/j.1756-8765.2010.01106.x},
	language     = {en}
}

@inproceedings{behnamghader2024llm2vec,
	title        = {{LLM2Vec: Large Language Models Are Secretly Powerful Text Encoders}},
	author       = {BehnamGhader, Parishad and Adlakha, Vaibhav and Mosbach, Marius and Bahdanau, Dzmitry and Chapados, Nicolas and Reddy, Siva},
	year         = 2024,
	booktitle={First Conference on Language Modeling}
}

@article{pezzelle2021word,
	title        = {{Word Representation Learning in Multimodal Pre-Trained Transformers: An Intrinsic Evaluation}},
	author       = {Pezzelle, Sandro and Takmaz, Ece and Fern{\'a}ndez, Raquel},
	year         = 2021,
	journal      = {Transactions of the Association for Computational Linguistics},
	publisher    = {MIT Press One Rogers Street, Cambridge, MA 02142-1209, USA journals-info~…},
	volume       = 9,
	pages        = {1563--1579}
}

@inproceedings{radford2021learning,
	title        = {{Learning Transferable Visual Models From Natural Language Supervision}},
	author       = {Radford, Alec and Kim, Jong Wook and Hallacy, Chris and Ramesh, Aditya and Goh, Gabriel and Agarwal, Sandhini and Sastry, Girish and Askell, Amanda and Mishkin, Pamela and Clark, Jack and others},
	year         = 2021,
	booktitle    = {International conference on machine learning},
	pages        = {8748--8763},
	organization = {PMLR}
}

@inproceedings{jia2021scaling,
	title        = {{Scaling Up Visual and Vision-Language Representation Learning With Noisy Text Supervision}},
	author       = {Jia, Chao and Yang, Yinfei and Xia, Ye and Chen, Yi-Ting and Parekh, Zarana and Pham, Hieu and Le, Quoc and Sung, Yun-Hsuan and Li, Zhen and Duerig, Tom},
	year         = 2021,
	booktitle    = {International conference on machine learning},
	pages        = {4904--4916},
	organization = {PMLR}
}

@article{li2019visualbert,
	title        = {{VisualBERT: A simple and Performant Baseline For Vision and Language}},
	author       = {Li, Liunian Harold and Yatskar, Mark and Yin, Da and Hsieh, Cho-Jui and Chang, Kai-Wei},
	year         = 2019,
	journal      = {arXiv preprint arXiv:1908.03557}
}

@article{kaup2024modal,
	title        = {Modal and amodal cognition: {A}n overarching principle in various domains of psychology},
	author       = {Kaup, Barbara and Ulrich, Rolf and Bausenhart, Karin M and Bryce, Donna and Butz, Martin V and Dignath, David and Dudschig, Carolin and Franz, Volker H and Friedrich, Claudia and Gawrilow, Caterina and others},
	year         = 2024,
	journal      = {Psychological Research},
	publisher    = {Springer},
	volume       = 88,
	number       = 2,
	pages        = {307--337}
}

@article{devereux2013representational,
	title        = {{Representational Similarity Analysis Reveals Commonalities and Differences in the Semantic Processing of Words and Objects}},
	author       = {Devereux, Barry J and Clarke, Alex and Marouchos, Andreas and Tyler, Lorraine K},
	year         = 2013,
	journal      = {Journal of Neuroscience},
	publisher    = {Soc Neuroscience},
	volume       = 33,
	number       = 48,
	pages        = {18906--18916}
}

@article{fairhall2013brain,
	title        = {{Brain Regions that Represent Amodal Conceptual Knowledge}},
	author       = {Fairhall, Scott L and Caramazza, Alfonso},
	year         = 2013,
	journal      = {Journal of Neuroscience},
	publisher    = {Soc Neuroscience},
	volume       = 33,
	number       = 25,
	pages        = {10552--10558}
}

@article{popham2021visual,
	title        = {{Visual and Linguistic Semantic Representations Are Aligned at the Border of Human Visual Cortex}},
	author       = {Popham, Sara F and Huth, Alexander G and Bilenko, Natalia Y and Deniz, Fatma and Gao, James S and Nunez-Elizalde, Anwar O and Gallant, Jack L},
	year         = 2021,
	journal      = {Nature neuroscience},
	publisher    = {Nature Publishing Group US New York},
	volume       = 24,
	number       = 11,
	pages        = {1628--1636}
}

@article{kriegeskorte2008representational,
	title        = {Representational similarity analysis-connecting the branches of systems neuroscience},
	author       = {Kriegeskorte, Nikolaus and Mur, Marieke and Bandettini, Peter A},
	year         = 2008,
	journal      = {Frontiers in systems neuroscience},
	publisher    = {Frontiers},
	volume       = 2,
	pages        = 249
}

@article{pereira2018toward,
	title        = {Toward a universal decoder of linguistic meaning from brain activation},
	author       = {Pereira, Francisco and Lou, Bin and Pritchett, Brianna and Ritter, Samuel and Gershman, Samuel J. and Kanwisher, Nancy and Botvinick, Matthew and Fedorenko, Evelina},
	year         = 2018,
	journal      = {Nature Communications},
	publisher    = {Nature Publishing Group UK London},
	volume       = 9,
	number       = 1,
	pages        = 963
}

@inproceedings{oota-etal-2022-neural,
	title        = {Neural Language Taskonomy: Which {NLP} Tasks are the most Predictive of f{MRI} Brain Activity?},
	author       = {Oota, Subba Reddy  and Arora, Jashn  and Agarwal, Veeral  and Marreddy, Mounika  and Gupta, Manish  and Surampudi, Bapi},
	year         = 2022,
	month        = jul,
	booktitle    = {Proceedings of the 2022 Conference of the North American Chapter of the Association for Computational Linguistics: Human Language Technologies},
	publisher    = {Association for Computational Linguistics},
	address      = {Seattle, United States},
	pages        = {3220--3237},
	doi          = {10.18653/v1/2022.naacl-main.235},
	url          = {https://aclanthology.org/2022.naacl-main.235},
	editor       = {Carpuat, Marine  and de Marneffe, Marie-Catherine  and Meza Ruiz, Ivan Vladimir}
}

@article{fedorenko2011functional,
	title        = {Functional specificity for high-level linguistic processing in the human brain},
	author       = {Fedorenko, Evelina and Behr, Michael K and Kanwisher, Nancy},
	year         = 2011,
	journal      = {Proceedings of the National Academy of Sciences},
	publisher    = {National Acad Sciences},
	volume       = 108,
	number       = 39,
	pages        = {16428--16433}
}

@article{buckner2008brain,
	title        = {The brain's default network: anatomy, function, and relevance to disease},
	author       = {Buckner, Randy L and Andrews-Hanna, Jessica R and Schacter, Daniel L},
	year         = 2008,
	journal      = {Annals of the new York Academy of Sciences},
	publisher    = {Wiley Online Library},
	volume       = 1124,
	number       = 1,
	pages        = {1--38}
}

@article{power2011functional,
	title        = {Functional network organization of the human brain},
	author       = {Power, Jonathan D and Cohen, Alexander L and Nelson, Steven M and Wig, Gagan S and Barnes, Kelly Anne and Church, Jessica A and Vogel, Alecia C and Laumann, Timothy O and Miezin, Fran M and Schlaggar, Bradley L and others},
	year         = 2011,
	journal      = {Neuron},
	publisher    = {Elsevier},
	volume       = 72,
	number       = 4,
	pages        = {665--678}
}

@article{laurenccon2024matters,
  title={What matters when building vision-language models?},
  author={Lauren{\c{c}}on, Hugo and Tronchon, L{\'e}o and Cord, Matthieu and Sanh, Victor},
  journal={arXiv preprint arXiv:2405.02246},
  year={2024}
}

@misc{liu2024llavanext,
    title={LLaVA-NeXT: Improved reasoning, OCR, and world knowledge},
    url={https://llava-vl.github.io/blog/2024-01-30-llava-next/},
    author={Liu, Haotian and Li, Chunyuan and Li, Yuheng and Li, Bo and Zhang, Yuanhan and Shen, Sheng and Lee, Yong Jae},
    month={January},
    year={2024}
}

@inproceedings{tan2019lxmert,
	title        = {LXMERT: Learning Cross-Modality Encoder Representations from Transformers},
	author       = {Tan, Hao and Bansal, Mohit},
	year         = 2019,
	booktitle    = {Proceedings of the 2019 Conference on Empirical Methods in Natural Language Processing and the 9th International Joint Conference on Natural Language Processing (EMNLP-IJCNLP)},
	pages        = {5100--5111}
}

@inproceedings{pennington2014glove,
	title        = {Glove: Global vectors for word representation},
	author       = {Pennington, Jeffrey and Socher, Richard and Manning, Christopher D},
	year         = 2014,
	booktitle    = {Proceedings of the 2014 conference on empirical methods in natural language processing (EMNLP)},
	pages        = {1532--1543}
}

@article{harnad1990symbol,
	title        = {The symbol grounding problem},
	author       = {Harnad, Stevan},
	year         = 1990,
	journal      = {Physica D: Nonlinear Phenomena},
	publisher    = {Elsevier},
	volume       = 42,
	number       = {1-3},
	pages        = {335--346}
}

@article{barsalou1999perceptual,
	title        = {Perceptual symbol systems},
	author       = {Barsalou, Lawrence W},
	year         = 1999,
	journal      = {Behavioral and brain sciences},
	publisher    = {Cambridge University Press},
	volume       = 22,
	number       = 4,
	pages        = {577--660}
}

@book{bergen2012louder,
	title        = {Louder than words: The new science of how the mind makes meaning},
	author       = {Bergen, Benjamin K},
	year         = 2012,
	publisher    = {Basic Books}
}

@article{simanova_modality-independent_2014,
	title        = {Modality-{Independent} {Decoding} of {Semantic} {Information} from the {Human} {Brain}},
	author       = {Simanova, Irina and Hagoort, Peter and Oostenveld, Robert and van Gerven, Marcel A. J.},
	year         = 2014,
	journal      = {Cerebral Cortex},
	volume       = 24,
	number       = 2,
	pages        = {426--434},
	doi          = {10.1093/cercor/bhs324},
	issn         = {1047-3211},
	url          = {https://doi.org/10.1093/cercor/bhs324},
	urldate      = {2023-10-24}
}

@inproceedings{
muennighoff2025generative,
title={Generative Representational Instruction Tuning},
author={Niklas Muennighoff and Hongjin SU and Liang Wang and Nan Yang and Furu Wei and Tao Yu and Amanpreet Singh and Douwe Kiela},
booktitle={The Thirteenth International Conference on Learning Representations},
year={2025},
url={https://openreview.net/forum?id=BC4lIvfSzv}
}

@article{Springer2024RepetitionIL,
  title={Repetition Improves Language Model Embeddings},
  author={Jacob M. Springer and Suhas Kotha and Daniel Fried and Graham Neubig and Aditi Raghunathan},
  journal={ArXiv},
  year={2024},
  volume={abs/2402.15449},
  url={https://api.semanticscholar.org/CorpusID:267897400}
}

@article{jiang2023mistral,
  title={Mistral 7B},
  author={Jiang, Albert Q and Sablayrolles, Alexandre and Mensch, Arthur and Bamford, Chris and Chaplot, Devendra Singh and Casas, Diego de las and Bressand, Florian and Lengyel, Gianna and Lample, Guillaume and Saulnier, Lucile and others},
  journal={arXiv preprint arXiv:2310.06825},
  year={2023}
}

@article{llama3,
  title={The {L}lama 3 herd of models},
  author={Grattafiori, Aaron and Dubey, Abhimanyu and Jauhri, Abhinav and Pandey, Abhinav and Kadian, Abhishek and Al-Dahle, Ahmad and Letman, Aiesha and Mathur, Akhil and Schelten, Alan and Vaughan, Alex and others},
  journal={arXiv preprint arXiv:2407.21783},
  year={2024}
}

@inproceedings{nikolaus2024modality,
  title={Modality-Agnostic fMRI Decoding of Vision and Language},
  author={Nikolaus, Mitja and Mozafari, Milad and Asher, Nicholas and Reddy, Leila and VanRullen, Rufin},
  booktitle={ICLR 2024 workshop on Representational Alignment (Re-Align)},
  year={2024}
}

@article{dong2023vision,
  title={Vision-Language Integration in Multimodal Video Transformers (Partially) Aligns with the Brain},
  author={Dong, Dota Tianai and Toneva, Mariya},
  journal={arXiv preprint arXiv:2311.07766},
  year={2023}
}

@inproceedings{devlin-etal-2019-bert,
    title = "{BERT}: Pre-training of Deep Bidirectional Transformers for Language Understanding",
    author = "Devlin, Jacob  and
      Chang, Ming-Wei  and
      Lee, Kenton  and
      Toutanova, Kristina",
    editor = "Burstein, Jill  and
      Doran, Christy  and
      Solorio, Thamar",
    booktitle = "Proceedings of the 2019 Conference of the North {A}merican Chapter of the Association for Computational Linguistics: Human Language Technologies, Volume 1 (Long and Short Papers)",
    month = jun,
    year = "2019",
    address = "Minneapolis, Minnesota",
    publisher = "Association for Computational Linguistics",
    url = "https://aclanthology.org/N19-1423/",
    doi = "10.18653/v1/N19-1423",
    pages = "4171--4186"
}

@article{xu2005language,
  title={Language in context: {E}mergent features of word, sentence, and narrative comprehension},
  author={Xu, Jiang and Kemeny, Stefan and Park, Grace and Frattali, Carol and Braun, Allen},
  journal={Neuroimage},
  volume={25},
  number={3},
  pages={1002--1015},
  year={2005},
  publisher={Elsevier}
}

@article{deniz2023semantic,
  title={Semantic representations during language comprehension are affected by context},
  author={Deniz, Fatma and Tseng, Christine and Wehbe, Leila and la Tour, Tom Dupr{\'e} and Gallant, Jack L},
  journal={Journal of Neuroscience},
  volume={43},
  number={17},
  pages={3144--3158},
  year={2023},
  publisher={Society for Neuroscience}
}

@InProceedings{winoground,
    author    = {Thrush, Tristan and Jiang, Ryan and Bartolo, Max and Singh, Amanpreet and Williams, Adina and Kiela, Douwe and Ross, Candace},
    title     = {{Winoground: Probing Vision and Language Models for Visio-Linguistic Compositionality}},
    booktitle = {Proceedings of the IEEE/CVF Conference on Computer Vision and Pattern Recognition (CVPR)},
    month     = {June},
    year      = {2022},
    pages     = {5238-5248}
}

@article{vsr,
    title = "{Visual Spatial Reasoning}",
    author = "Liu, Fangyu  and
      Emerson, Guy  and
      Collier, Nigel",
    journal = "Transactions of the Association for Computational Linguistics",
    volume = "11",
    year = "2023",
    address = "Cambridge, MA",
    publisher = "MIT Press",
    url = "https://aclanthology.org/2023.tacl-1.37",
    doi = "10.1162/tacl_a_00566",
    pages = "635--651",
}

@inproceedings{svo_probing,
    title = "{Probing Image-Language Transformers for Verb Understanding}",
    author = "Hendricks, Lisa Anne  and
      Nematzadeh, Aida",
    editor = "Zong, Chengqing  and
      Xia, Fei  and
      Li, Wenjie  and
      Navigli, Roberto",
    booktitle = "Findings of the Association for Computational Linguistics: ACL-IJCNLP 2021",
    month = aug,
    year = "2021",
    address = "Online",
    publisher = "Association for Computational Linguistics",
    url = "https://aclanthology.org/2021.findings-acl.318",
    doi = "10.18653/v1/2021.findings-acl.318",
    pages = "3635--3644",
}

@inproceedings{valse,
    title = "{VALSE: A Task-Independent Benchmark for Vision and Language Models Centered on Linguistic Phenomena}",
    author = "Parcalabescu, Letitia  and
      Cafagna, Michele  and
      Muradjan, Lilitta  and
      Frank, Anette  and
      Calixto, Iacer  and
      Gatt, Albert",
    editor = "Muresan, Smaranda  and
      Nakov, Preslav  and
      Villavicencio, Aline",
    booktitle = "Proceedings of the 60th Annual Meeting of the Association for Computational Linguistics (Volume 1: Long Papers)",
    month = may,
    year = "2022",
    address = "Dublin, Ireland",
    publisher = "Association for Computational Linguistics",
    url = "https://aclanthology.org/2022.acl-long.567",
    doi = "10.18653/v1/2022.acl-long.567",
    pages = "8253--8280",
}

@inproceedings{
ryskina2025language,
title={Language models align with brain regions that represent concepts across modalities},
author={Maria Ryskina and Greta Tuckute and Alexander Fung and Ashley Malkin and Evelina Fedorenko},
booktitle={Second Conference on Language Modeling},
year={2025},
url={https://openreview.net/forum?id=2JohTFaGbW}
}

@InProceedings{Zhai_2022_CVPR,
    author    = {Zhai, Xiaohua and Wang, Xiao and Mustafa, Basil and Steiner, Andreas and Keysers, Daniel and Kolesnikov, Alexander and Beyer, Lucas},
    title     = {LiT: Zero-Shot Transfer With Locked-Image Text Tuning},
    booktitle = {Proceedings of the IEEE/CVF Conference on Computer Vision and Pattern Recognition (CVPR)},
    month     = {June},
    year      = {2022},
    pages     = {18123-18133}
}

@article{deitke2024molmo,
  title={Molmo and {P}ix{M}o: Open Weights and Open Data for State-of-the-Art Multimodal Models},
  author={Deitke, Matt and Clark, Christopher and Lee, Sangho and Tripathi, Rohun and Yang, Yue and Park, Jae Sung and Salehi, Mohammadreza and Muennighoff, Niklas and Lo, Kyle and Soldaini, Luca and others},
  journal={CoRR},
  year={2024}
}

@article{bai2025qwen2,
  title={Qwen2.5-{VL} technical report},
  author={Bai, Shuai and Chen, Keqin and Liu, Xuejing and Wang, Jialin and Ge, Wenbin and Song, Sibo and Dang, Kai and Wang, Peng and Wang, Shijie and Tang, Jun and others},
  journal={arXiv preprint arXiv:2502.13923},
  year={2025}
}

@inproceedings{bavaresco-fernandez-2025-experiential,
    title = "Experiential Semantic Information and Brain Alignment: Are Multimodal Models Better than Language Models?",
    author = "Bavaresco, Anna  and
      Fern{\'a}ndez, Raquel",
    editor = "Boleda, Gemma  and
      Roth, Michael",
    booktitle = "Proceedings of the 29th Conference on Computational Natural Language Learning",
    month = jul,
    year = "2025",
    address = "Vienna, Austria",
    publisher = "Association for Computational Linguistics",
    url = "https://aclanthology.org/2025.conll-1.10/",
    doi = "10.18653/v1/2025.conll-1.10",
    pages = "141--155",
    ISBN = "979-8-89176-271-8"
}

@inproceedings{zhuang-etal-2024-visual,
    title = "Visual Grounding Helps Learn Word Meanings in Low-Data Regimes",
    author = "Zhuang, Chengxu  and
      Fedorenko, Evelina  and
      Andreas, Jacob",
    editor = "Duh, Kevin  and
      Gomez, Helena  and
      Bethard, Steven",
    booktitle = "Proceedings of the 2024 Conference of the North American Chapter of the Association for Computational Linguistics: Human Language Technologies (Volume 1: Long Papers)",
    month = jun,
    year = "2024",
    address = "Mexico City, Mexico",
    publisher = "Association for Computational Linguistics",
    url = "https://aclanthology.org/2024.naacl-long.71/",
    doi = "10.18653/v1/2024.naacl-long.71",
    pages = "1311--1329"
}

@article{zwaan2014embodiment,
  title={Embodiment and language comprehension: Reframing the discussion},
  author={Zwaan, Rolf A},
  journal={Trends in cognitive sciences},
  volume={18},
  number={5},
  pages={229--234},
  year={2014},
  publisher={Elsevier}
}

@article{handjaras2016concepts,
  title={How concepts are encoded in the human brain: a modality independent, category-based cortical organization of semantic knowledge},
  author={Handjaras, Giacomo and Ricciardi, Emiliano and Leo, Andrea and Lenci, Alessandro and Cecchetti, Luca and Cosottini, Mirco and Marotta, Giovanna and Pietrini, Pietro},
  journal={Neuroimage},
  volume={135},
  pages={232--242},
  year={2016},
  publisher={Elsevier}
}

@article{bonner2013heteromodal,
  title={Heteromodal conceptual processing in the angular gyrus},
  author={Bonner, Michael F and Peelle, Jonathan E and Cook, Philip A and Grossman, Murray},
  journal={Neuroimage},
  volume={71},
  pages={175--186},
  year={2013},
  publisher={Elsevier}
}

@article{shepard1970second,
  title={Second-order isomorphism of internal representations: {S}hapes of states},
  author={Shepard, Roger N and Chipman, Susan},
  journal={Cognitive psychology},
  volume={1},
  number={1},
  pages={1--17},
  year={1970},
  publisher={Elsevier}
}

@inproceedings{dujmovic2024inferring,
  title={Inferring {DNN}-Brain Alignment using Representational Similarity Analyses can be Problematic},
  author={Dujmovic, Marin and Bowers, Jeffrey and Adolfi, Federico and Malhotra, Gaurav},
  booktitle={ICLR 2024 Workshop on Representational Alignment},
  year={2024}
}

@article{ivanova2021language,
  title={The language network is recruited but not required for nonverbal event semantics},
  author={Ivanova, Anna A and Mineroff, Zachary and Zimmerer, Vitor and Kanwisher, Nancy and Varley, Rosemary and Fedorenko, Evelina},
  journal={Neurobiology of Language},
  volume={2},
  number={2},
  pages={176--201},
  year={2021},
  publisher={MIT Press One Rogers Street, Cambridge, MA 02142-1209, USA journals-info~…}
}

@article{naselaris2011encoding,
  title={Encoding and decoding in f{MRI}},
  author={Naselaris, Thomas and Kay, Kendrick N and Nishimoto, Shinji and Gallant, Jack L},
  journal={Neuroimage},
  volume={56},
  number={2},
  pages={400--410},
  year={2011},
  publisher={Elsevier}
}

@InProceedings{Madasu_2023_CVPR,
    author    = {Madasu, Avinash and Lal, Vasudev},
    title     = {Is Multimodal Vision Supervision Beneficial to Language?},
    booktitle = {Proceedings of the IEEE/CVF Conference on Computer Vision and Pattern Recognition (CVPR) Workshops},
    month     = {June},
    year      = {2023},
    pages     = {2637-2642}
}

@inproceedings{iki-aizawa-2021-effect,
    title = "Effect of Visual Extensions on Natural Language Understanding in Vision-and-Language Models",
    author = "Iki, Taichi  and
      Aizawa, Akiko",
    editor = "Moens, Marie-Francine  and
      Huang, Xuanjing  and
      Specia, Lucia  and
      Yih, Scott Wen-tau",
    booktitle = "Proceedings of the 2021 Conference on Empirical Methods in Natural Language Processing",
    month = nov,
    year = "2021",
    address = "Online and Punta Cana, Dominican Republic",
    publisher = "Association for Computational Linguistics",
    url = "https://aclanthology.org/2021.emnlp-main.167/",
    doi = "10.18653/v1/2021.emnlp-main.167",
    pages = "2189--2196"
}

@inproceedings{
qin2025visionandlanguage,
title={Vision-and-Language Training Helps Deploy Taxonomic Knowledge but Does Not Fundamentally Alter It},
author={Yulu Qin and Dheeraj Varghese and Adam Dahlgren Lindstr{\"o}m and Lucia Donatelli and Kanishka Misra and Najoung Kim},
booktitle={The Thirty-ninth Annual Conference on Neural Information Processing Systems},
year={2025},
url={https://openreview.net/forum?id=KXmDTGKwhy}
}

@article{wang2023finding,
  title={Finding structure in one child's linguistic experience},
  author={Wang, Wentao and Vong, Wai Keen and Kim, Najoung and Lake, Brenden M},
  journal={Cognitive science},
  volume={47},
  number={6},
  pages={e13305},
  year={2023},
  publisher={Wiley Online Library}
}

\appendix

\section{Data}

\subsection{Concepts}
\label{sec:appendix_pereira_dataset}

The full list of concept words from the Pereira dataset \citep{pereira2018toward} is the following:
\begin{quote}

Ability,
Accomplished,
Angry,
Apartment,
Applause,
Argument,
Argumentatively,
Art,
Attitude,
Bag,
Ball,
Bar,
Bear,
Beat,
Bed,
Beer,
Big,
Bird,
Blood,
Body,
Brain,
Broken,
Building,
Burn,
Business,
Camera,
Carefully,
Challenge,
Charity,
Charming,
Clothes,
Cockroach,
Code,
Collection,
Computer,
Construction,
Cook,
Counting,
Crazy,
Damage,
Dance,
Dangerous,
Deceive,
Dedication,
Deliberately,
Delivery,
Dessert,
Device,
Dig,
Dinner,
Disease,
Dissolve,
Disturb,
Do,
Doctor,
Dog,
Dressing,
Driver,
Economy,
Election,
Electron,
Elegance,
Emotion,
Emotionally,
Engine,
Event,
Experiment,
Extremely,
Feeling,
Fight,
Fish,
Flow,
Food,
Garbage,
Gold,
Great,
Gun,
Hair,
Help,
Hurting,
Ignorance,
Illness,
Impress,
Invention,
Investigation,
Invisible,
Job,
Jungle,
Kindness,
King,
Lady,
Land,
Laugh,
Law,
Left,
Level,
Liar,
Light,
Magic,
Marriage,
Material,
Mathematical,
Mechanism,
Medication,
Money,
Mountain,
Movement,
Movie,
Music,
Nation,
News,
Noise,
Obligation,
Pain,
Personality,
Philosophy,
Picture,
Pig,
Plan,
Plant,
Play,
Pleasure,
Poor,
Prison,
Professional,
Protection,
Quality,
Reaction,
Read,
Relationship,
Religious,
Residence,
Road,
Sad,
Science,
Seafood,
Sell,
Sew,
Sexy,
Shape,
Ship,
Show,
Sign,
Silly,
Sin,
Skin,
Smart,
Smiling,
Solution,
Soul,
Sound,
Spoke,
Star,
Student,
Stupid,
Successful,
Sugar,
Suspect,
Table,
Taste,
Team,
Texture,
Time,
Tool,
Toy,
Tree,
Trial,
Tried,
Typical,
Unaware,
Usable,
Useless,
Vacation,
War,
Wash,
Weak,
Wear,
Weather,
Willingly,
Word.
\end{quote}

\subsection{fMRI responses and preprocessing}
\label{sec:brain_data}

\citet{pereira2018toward} preprocessed the fMRI data by estimating the response 
to each stimulus using a general linear model in which each stimulus presentation 
(sentence or word and picture) was modeled with a boxcar function convolved with 
the canonical hemodynamic response; the
voxelwise beta estimates are what we referred to as \textit{brain responses} or \textit{vixel-wise activations} throughout the paper.

\begin{table*}\centering \small
\begin{tabular}{lll}\toprule
\bf Network & \multicolumn{2}{c}{\bf Regions of Interest}\\ \midrule
\textit{Language} 
            & - Posterior temporal lobe & - Inferior frontal gyrus\\
            & - Anterior temporal lobe & - Middle frontal gyrus\\
            & - Angular gyrus & - Orbitan inferior frontal gyrus \\ \midrule
\textit{Visual}   
            & - Parahippocampal place area & - Superior temporal sulcus\\
            & - Retrosplenial cortex & - Fusiform face area \\
            & - Transverse occipital sulcus & - Occipital face area\\
            & - Lateral occipital area  & - Extrastriate body area \\
            \bottomrule
\end{tabular}
\caption{Brain regions of interest (ROIs) corresponding to our investigated functionally localised \textit{language} and \textit{visual} networks.}
\label{tab:brain_areas_overview}
\end{table*}

\section{Extended Methods}

\subsection{Model details}
\label{sec:model_details}
We use both VLMs and their language-only counterparts off-the-shelf, as publicly available in the HuggingFace\footnote{\url{https://huggingface.co/}} library. Below, we report the HuggingFace IDs of the specific model implementations used in our experiments:
\begin{itemize}
    \item \texttt{openai/clip-vit-large-patch14}
    \item \texttt{kakaobrain/align-base}
    \item \texttt{unc-nlp/lxmert-base-uncased}
    \item \texttt{uclanlp/visualbert-vqa-coco-pre}
    \item \texttt{HuggingFaceM4/idefics2-8b}
    \item \texttt{llava-hf/llama3-llava-next-8b-hf}
    \item \texttt{bert-base-uncased}
    \item \texttt{mistralai/Mistral-7B-v0.1}
    \item \texttt{meta-llama/Meta-Llama-3-8B-Instruct}
\end{itemize}

As for the GloVe representations included in our study, they were extracted using the official vectors\footnote{\url{https://nlp.stanford.edu/projects/glove/}} pretrained on CommonCrawl (840B tokens version).
\subsection{Partial correlation analysis}
\label{sec:partial_correlations}
\begin{figure}
    \centering
    \includegraphics[width=0.95\linewidth]{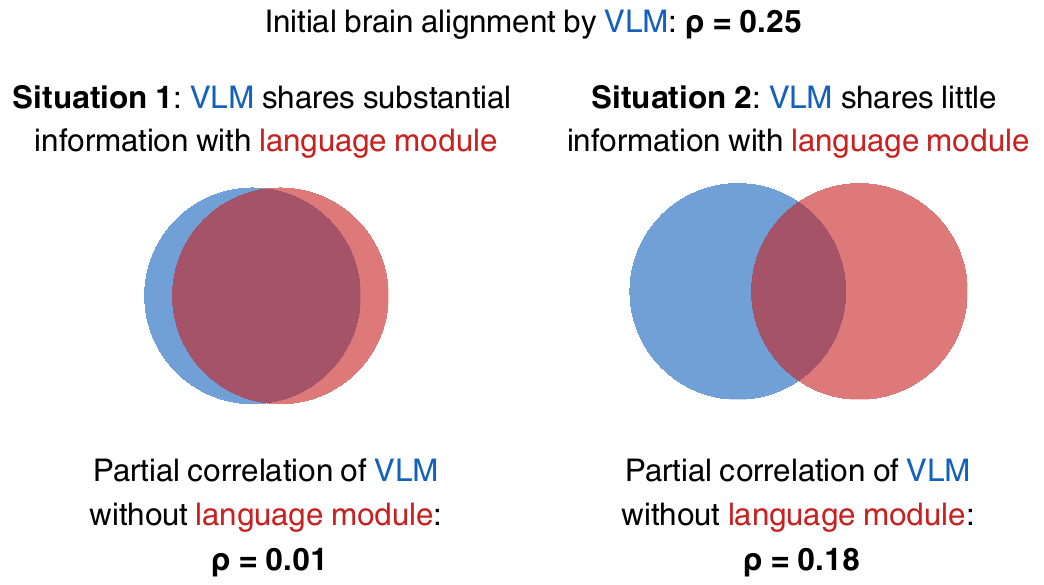}
    \caption{Schematic illustrating situations that can be disambiguated by computing partial correlations. If the initial brain alignment of a VLM is attributable to information substantially shared with the language-only module, the partial correlation will be significantly weaker than the initial correlation.}
    \label{fig:venn-diagram}
\end{figure}

To better illustrate how partial correlation analysis can help disentangle the representational-alignment contributions by different models, we provide a visualisation in Figure~\ref{fig:venn-diagram}.

\section{Detailed Results}

\subsection{Comparisons between models}
\label{sec:model_comparisons}

Table~\ref{tab:p_vals} provides all the `raw' (i.e., not yet Bonferroni corrected) $p$-values we calculate in the main experiment to assess whether pairwise differences between models are statistically significant.

\begin{table*}[]
    \centering
    \begin{tabular}{llllll} \toprule
       Model Pair  &  \multicolumn{2}{c}{LH Language} & \multicolumn{2}{c}{Visual}\\ 
       & \textit{Sent.} & \textit{Pic.} & \textit{Sent.} & \textit{Pic.}  \\\midrule
       CLIP - ALIGN & 0.001 & 0.014 & 0.051 & 0.0  \\
        CLIP - LXMERT & 0.0 & 0.0 & 0.0 & 0.229  \\
        CLIP - VisualBERT & 0.0 & 0.0 & 0.0 & 0.0  \\
        CLIP - IDEFICS2 & 0.745 & 0.0 & 0.0 & 0.0  \\
        CLIP - LLaVA & 0.0 & 0.0 & 0.013 & 0.0  \\
        CLIP - Mistral & 0.567 & 0.0 & 0.0 & 0.0  \\
        CLIP - Llama3 & 0.0 & 0.0 & 0.555 & 0.0  \\
        CLIP - BERT & 0.001 & 0.0 & 0.234 & 0.0  \\
        CLIP - GloVe & 0.097 & 0.0 &0.001 & 0.0  \\
        ALIGN - LXMERT & 0.0 & 0.004 & 0.0 & 0.0  \\
        ALIGN - VisualBERT & 0.0 & 0.005 & 0.0 & 0.0  \\
        ALIGN - IDEFICS2 & 0.002 & 0.0 & 0.0 & 0.0  \\
        ALIGN - LLaVA & 0.324 & 0.0 & 0.0 & 0.0  \\
        ALIGN - Mistral & 0.004 & 0.0 & 0.0 & 0.0  \\
        ALIGN - Llama3 & 0.078 & 0.0 & 0.011 & 0.0  \\
        ALIGN - BERT & 0.823 & 0.0 & 0.447 & 0.0  \\
        ALIGN - GloVe & 0.0 & 0.0 &  0.147 & 0.0  \\
        LXMERT - VisualBERT & 0.155 & 0.977 & 0.037 & 0.0  \\
        LXMERT - IDEFICS2 & 0.0 & 0.0 & 0.135 & 0.0  \\
        LXMERT - LLaVA & 0.0 & 0.0 & 0.033 & 0.0  \\
        LXMERT - Mistral & 0.0 & 0.0 & 0.982 & 0.0  \\
        LXMERT - Llama3 & 0.0 & 0.0 & 0.0 & 0.0  \\
        LXMERT - BERT & 0.0 & 0.0 & 0.0 & 0.0  \\
        LXMERT - GloVe & 0.008 & 0.0 & 0.0 & 0.0  \\
        VisualBERT - IDEFICS2 & 0.0 & 0.0 & 0.552 & 0.0  \\
        VisualBERT - LLaVA & 0.0 & 0.0 & 0.0 & 0.0  \\
        VisualBERT - Mistral & 0.0 & 0.0 & 0.035 & 0.0  \\
        VisualBERT - Llama3 & 0.0 & 0.0 & 0.0 & 0.0  \\
        VisualBERT - BERT & 0.0 & 0.0 & 0.0 & 0.0  \\
        VisualBERT - GloVe & 0.0 & 0.0 & 0.0 & 0.0  \\
        IDEFICS2 - LLaVA & 0.0 & 0.01 & 0.0 & 0.0  \\
        IDEFICS2 - Mistral & 0.805 & 0.0 & 0.129 & 0.0  \\
        IDEFICS2 - Llama3 & 0.0 & 0.0 & 0.0 & 0.0  \\
        IDEFICS2 - BERT & 0.004 & 0.0 & 0.0 & 0.0  \\
        IDEFICS2 - GloVe & 0.047 & 0.0 & 0.0 & 0.063  \\
        LLaVA - Mistral & 0.0 & 0.0 & 0.035 & 0.0  \\
        LLaVA - Llama3 & 0.438 & 0.0 & 0.059 & 0.0  \\
        LLaVA - BERT & 0.226 & 0.0 & 0.0 & 0.0  \\
        LLaVA - GloVe & 0.0 & 0.0 & 0.0 & 0.001  \\
        Mistral - Llama3 & 0.0 & 0.0 & 0.0 & 0.001  \\
        Mistral - BERT & 0.008 & 0.0 & 0.0 & 0.002  \\
        Mistral - GloVe & 0.026 & 0.0 & 0.0 & 0.0  \\
        Llama3 - BERT & 0.047 & 0.0 & 0.075 & 0.819  \\
        Llama3 - GloVe & 0.0 & 0.0 & 0.0 & 0.0  \\
        BERT - GloVe & 0.0 & 0.072 & 0.027 & 0.0  \\ \bottomrule
    \end{tabular}
    \caption{$p$-values associated with all pairwise model comparisons in the main experiment before Bonferroni corrections.}
    \label{tab:p_vals}
\end{table*}

\subsection{Layer-wise RSA results}
\label{sec:appendix_layerwise_results}

We provide layer-wise RSA results for the sentence condition in Tables~\ref{tab:language_LH_small_models}, \ref{tab:language_LH_large_models}, \ref{tab:visual_small_models} and \ref{tab:visual_large_models}. Layer-wise results for the picture condition are reported in Tables~\ref{tab:language_LH_large_models_picture} and \ref{tab:visual_large_models_picture}. In each of these tables, we boldface the highest correlation value (which was identified before rounding to the second decimal) and check whether it is statistically significantly different from the correlation achieved by each of the other layers by performing pairwise comparisons as described in the main paper. $p$-values are Bonferroni-corrected, with the number of comparisons amounting to $1 - \#layers$ for each model.

\begin{table*}
    \centering
    \begin{tabular}{llllllllll} \toprule
        Layer & CLIP & ALIGN & LXMERT & VisualBERT & BERT \\ \midrule
        1 & -0.06 (0.0)* & -0.0 (0.58)* & 0.07 (0.0) & \textbf{0.1} (0.0) & 0.01 (0.13) \\
        2 & 0.05 (0.0) & -0.01 (0.22)* & 0.07 (0.0) & 0.08 (0.0) & \textbf{0.02} (0.0) \\
        3 & 0.03 (0.0) & 0.01 (0.13) & 0.07 (0.0) & 0.07 (0.0) & 0.02 (0.0) \\
        4 & 0.02 (0.0)* & \textbf{0.02} (0.01) & \textbf{0.08 (0.0)} & 0.07 (0.0) & 0.02 (0.01) \\
        5 & 0.01 (0.43)* & 0.01 (0.2) & 0.06 (0.0) & 0.07 (0.0) & 0.02 (0.03) \\
        6 & -0.0 (0.64)* & -0.0 (0.9) & 0.05 (0.0)* & 0.07 (0.0)* & 0.02 (0.01) \\
        7 & 0.01 (0.13)* & 0.0 (0.81) & 0.04 (0.0)* & 0.06 (0.0)* & 0.02 (0.05) \\
        8 & 0.0 (0.68)* & 0.0 (0.81) & 0.03 (0.0)* & 0.06 (0.0)* & 0.01 (0.3) \\
        9 & 0.01 (0.24)* & 0.0 (0.63) & 0.04 (0.0)* & 0.05 (0.0)* & -0.0 (0.58)* \\
        10 & -0.0 (0.66)* & 0.0 (0.74) & 0.04 (0.0)* & 0.03 (0.0)* & -0.03 (0.0)* \\
        11 & 0.01 (0.12)* & -0.0 (0.97) & 0.03 (0.0)* & 0.01 (0.22)* & -0.05 (0.0)* \\
        12 & 0.01 (0.1)* & 0.0 (0.69) & -0.03 (0.0)* & -0.0 (0.57)* & -0.05 (0.0)* \\
        13 & \textbf{0.05} (0.0) & 0.02 (0.02) & -0.08 (0.0)* & 0.0 (0.55)* & -0.08 (0.0)* \\
        14 & - & - & 0.03 (0.0)* & - & - \\\bottomrule  
        \end{tabular}
    \caption{Layer-wise RSA values (Spearman correlations) for the sentence condition and LH language network, with $p$-values indicating the significance of the correlation (i.e., whether it is different from 0) in brackets. Note that asterisks indicate whether each correlation is statistically significantly different from the correlation achieved by the best layer (boldfaced value).}
    \label{tab:language_LH_small_models}
\end{table*}

\begin{table*}
    \centering
    
    \begin{tabular}{llllllllll} \toprule

        Layer & CLIP & ALIGN & LXMERT & VisualBERT & BERT \\ \midrule
        1 & -0.04 (0.0)* & 0.0 (0.76) & 0.06 (0.0) & \textbf{0.09} (0.0) & 0.02 (0.01) \\
        2 & \textbf{0.04} (0.0) & 0.0 (1.0) & 0.06 (0.0) & 0.07 (0.0) & 0.02 (0.01) \\
        3 & 0.03 (0.0) & 0.02 (0.05) & 0.06 (0.0) & 0.07 (0.0) & 0.02 (0.02) \\
        4 & 0.02 (0.0) & \textbf{0.02} (0.01) & \textbf{0.08} (0.0) & 0.08 (0.0) & \textbf{0.03} (0.0) \\
        5 & 0.01 (0.38)* & 0.01 (0.47) & 0.06 (0.0) & 0.08 (0.0) & 0.03 (0.0) \\
        6 & -0.01 (0.11)* & -0.0 (0.68)* & 0.05 (0.0)* & 0.07 (0.0) & 0.02 (0.01) \\
        7 & 0.01 (0.07)* & 0.0 (0.96) & 0.05 (0.0)* & 0.08 (0.0) & 0.02 (0.01) \\
        8 & 0.0 (0.85)* & -0.01 (0.33)* & 0.04 (0.0)* & 0.07 (0.0) & 0.02 (0.04) \\
        9 & 0.02 (0.01) & -0.01 (0.18)* & 0.03 (0.0)* & 0.05 (0.0)* & -0.0 (0.77)* \\
        10 & 0.01 (0.48)* & -0.02 (0.02)* & 0.04 (0.0)* & 0.04 (0.0)* & -0.02 (0.0)* \\
        11 & 0.02 (0.0) & -0.02 (0.04)* & 0.03 (0.0)* & 0.02 (0.03)* & -0.03 (0.0)* \\
        12 & 0.01 (0.1)* & -0.01 (0.08)* & -0.02 (0.01)* & 0.02 (0.04)* & -0.02 (0.0)* \\
        13 & 0.01 (0.5)* & -0.0 (0.66)* & -0.05 (0.0)* & 0.03 (0.0)* & -0.04 (0.0)* \\
        14 & - & - & 0.05 (0.0)* & - & - \\\bottomrule  
        \end{tabular}
    \caption{Layer-wise RSA values (Spearman correlations) for the sentence condition and visual network, with $p$-values indicating the significance of the correlation (i.e., whether it is different from 0) in brackets. Note that asterisks indicate whether each correlation is statistically significantly different from the correlation achieved by the best layer (boldfaced value).}
    \label{tab:visual_small_models}
\end{table*}

\begin{table*}
    \centering
    \begin{tabular}{lllllllll} \toprule
        Layer & LLaVA & IDEFICS2 & Llama3 & Mistral \\ \midrule
        1 & -0.05 (0.0)* & -0.01 (0.39)* & -0.05 (0.0)* & -0.01 (0.39)* \\
        2 & -0.02 (0.03)* & 0.02 (0.01)* & -0.03 (0.0)* & 0.02 (0.02)* \\
        3 & 0.0 (0.71) & -0.01 (0.23)* & -0.02 (0.0)* & -0.01 (0.48)* \\
        4 & 0.01 (0.5) & -0.03 (0.0)* & 0.0 (0.99) & -0.05 (0.0)* \\
        5 & -0.0 (0.83) & -0.03 (0.0)* & -0.0 (0.91) & -0.04 (0.0)* \\
        6 & -0.03 (0.0)* & 0.02 (0.05)* & -0.02 (0.03) & 0.01 (0.45)* \\
        7 & -0.05 (0.0)* & -0.01 (0.29)* & -0.04 (0.0)* & -0.02 (0.02)* \\
        8 & -0.05 (0.0)* & -0.03 (0.0)* & -0.04 (0.0)* & -0.02 (0.0)* \\
        9 & -0.07 (0.0)* & -0.06 (0.0)* & -0.06 (0.0)* & -0.04 (0.0)* \\
        10 & -0.07 (0.0)* & -0.07 (0.0)* & -0.07 (0.0)* & -0.05 (0.0)* \\
        11 & -0.08 (0.0)* & -0.06 (0.0)* & -0.08 (0.0)* & -0.05 (0.0)* \\
        12 & -0.07 (0.0)* & -0.05 (0.0)* & -0.08 (0.0)* & -0.05 (0.0)* \\
        13 & -0.08 (0.0)* & -0.05 (0.0)* & -0.09 (0.0)* & -0.05 (0.0)* \\
        14 & -0.09 (0.0)* & -0.06 (0.0)* & -0.09 (0.0)* & -0.06 (0.0)* \\
        15 & -0.09 (0.0)* & -0.04 (0.0)* & -0.09 (0.0)* & -0.04 (0.0)* \\
        16 & -0.06 (0.0)* & -0.04 (0.0)* & -0.04 (0.0)* & -0.04 (0.0)* \\
        17 & -0.05 (0.0)* & -0.02 (0.01)* & -0.01 (0.16) & -0.02 (0.01)* \\
        18 & -0.07 (0.0)* & -0.04 (0.0)* & -0.02 (0.01)* & -0.03 (0.0)* \\
        19 & -0.04 (0.0)* & -0.03 (0.0)* & 0.0 (0.81) & -0.03 (0.0)* \\
        20 & -0.04 (0.0)* & 0.0 (0.69)* & \textbf{0.01} (0.41) & 0.02 (0.04)* \\
        21 & -0.05 (0.0)* & 0.02 (0.0) & 0.0 (0.54) & 0.04 (0.0) \\
        22 & -0.04 (0.0)* & 0.03 (0.0) & -0.0 (0.63) & 0.04 (0.0) \\
        23 & -0.02 (0.02)* & 0.04 (0.0) & 0.0 (0.6) & \textbf{0.04} (0.0) \\
        24 & -0.02 (0.0)* & \textbf{0.05} (0.0) & -0.01 (0.18) & 0.04 (0.0) \\
        25 & -0.02 (0.01)* & 0.04 (0.0) & -0.02 (0.02) & 0.03 (0.0) \\
        26 & -0.02 (0.0)* & 0.03 (0.0) & -0.03 (0.0)* & 0.02 (0.01) \\
        27 & -0.03 (0.0)* & 0.03 (0.0) & -0.03 (0.0)* & 0.02 (0.03)* \\
        28 & -0.03 (0.0)* & 0.02 (0.0) & -0.03 (0.0)* & 0.01 (0.18)* \\
        29 & -0.03 (0.0)* & 0.01 (0.26)* & -0.04 (0.0)* & -0.0 (0.81)* \\
        30 & -0.02 (0.0)* & -0.01 (0.33)* & -0.03 (0.0)* & -0.01 (0.17)* \\
        31 & -0.03 (0.0)* & -0.02 (0.05)* & -0.04 (0.0)* & -0.01 (0.36)* \\
        32 & -0.02 (0.04)* & -0.02 (0.05)* & -0.04 (0.0)* & -0.01 (0.11)* \\
        33 & \textbf{0.01} (0.1) & -0.01 (0.52)* & 0.0 (0.57) & -0.01 (0.52)* \\ \bottomrule  
        \end{tabular}
    \caption{Layer-wise RSA values (Spearman correlations) for the sentence condition and LH language network, with $p$-values indicating the significance of the correlation (i.e., whether it is different from 0) in brackets. Note that asterisks indicate whether each correlation is statistically significantly different from the correlation achieved by the best layer (boldfaced value).}
    \label{tab:language_LH_large_models}
\end{table*}

\begin{table*}
    \centering
    \begin{tabular}{lllllllll} \toprule
        Layer & LLaVA & IDEFICS2 & Llama3 & Mistral \\ \midrule

        1 & -0.01 (0.08)* & 0.01 (0.27)* & -0.02 (0.05)* & 0.01 (0.27)* \\
        2 & -0.02 (0.04)* & -0.01 (0.2)* & -0.02 (0.01)* & 0.0 (0.75)* \\
        3 & 0.01 (0.07)* & 0.02 (0.0)* & -0.01 (0.47)* & 0.02 (0.01)* \\
        4 & 0.03 (0.0)* & -0.01 (0.28)* & 0.0 (0.95)* & -0.02 (0.06)* \\
        5 & 0.01 (0.11)* & -0.01 (0.08)* & 0.0 (0.71)* & -0.02 (0.0)* \\
        6 & 0.01 (0.07)* & 0.0 (0.93)* & 0.02 (0.02) & -0.0 (0.86)* \\
        7 & -0.0 (0.54)* & -0.0 (0.6)* & -0.0 (0.78)* & -0.01 (0.32)* \\
        8 & -0.01 (0.08)* & -0.02 (0.05)* & -0.02 (0.04)* & -0.01 (0.17)* \\
        9 & -0.02 (0.0)* & -0.03 (0.0)* & -0.03 (0.0)* & -0.01 (0.1)* \\
        10 & -0.03 (0.0)* & -0.05 (0.0)* & -0.04 (0.0)* & -0.03 (0.0)* \\
        11 & -0.03 (0.0)* & -0.05 (0.0)* & -0.05 (0.0)* & -0.03 (0.0)* \\
        12 & -0.03 (0.0)* & -0.03 (0.0)* & -0.05 (0.0)* & -0.03 (0.0)* \\
        13 & -0.03 (0.0)* & -0.03 (0.0)* & -0.05 (0.0)* & -0.03 (0.0)* \\
        14 & -0.06 (0.0)* & -0.03 (0.0)* & -0.06 (0.0)* & -0.03 (0.0)* \\
        15 & -0.06 (0.0)* & -0.03 (0.0)* & -0.06 (0.0)* & -0.03 (0.0)* \\
        16 & -0.02 (0.01)* & -0.04 (0.0)* & -0.01 (0.08)* & -0.04 (0.0)* \\
        17 & -0.01 (0.15)* & -0.02 (0.01)* & -0.0 (0.81)* & -0.02 (0.0)* \\
        18 & -0.03 (0.0)* & -0.03 (0.0)* & -0.02 (0.05)* & -0.02 (0.0)* \\
        19 & 0.0 (0.54)* & -0.03 (0.0)* & 0.01 (0.29)* & -0.03 (0.0)* \\
        20 & 0.03 (0.0)* & 0.03 (0.0)* & 0.02 (0.01) & 0.03 (0.0)* \\
        21 & 0.02 (0.01)* & 0.04 (0.0)* & 0.02 (0.0) & 0.04 (0.0)* \\
        22 & 0.03 (0.0) & 0.06 (0.0) & 0.03 (0.0) & 0.06 (0.0) \\
        23 & 0.05 (0.0) & 0.08 (0.0) & \textbf{0.04} (0.0) & 0.07 (0.0) \\
        24 & 0.06 (0.0) & 0.09 (0.0) & 0.04 (0.0) & \textbf{0.07} (0.0) \\
        25 & \textbf{0.06} (0.0) & \textbf{0.09} (0.0) & 0.03 (0.0) & 0.07 (0.0) \\
        26 & 0.05 (0.0) & 0.09 (0.0) & 0.02 (0.0) & 0.07 (0.0) \\
        27 & 0.05 (0.0) & 0.09 (0.0) & 0.02 (0.01) & 0.07 (0.0) \\
        28 & 0.05 (0.0) & 0.08 (0.0) & 0.02 (0.05)* & 0.06 (0.0) \\
        29 & 0.04 (0.0) & 0.06 (0.0) & 0.01 (0.15)* & 0.05 (0.0)* \\
        30 & 0.05 (0.0) & 0.06 (0.0)* & 0.01 (0.12)* & 0.05 (0.0) \\
        31 & 0.03 (0.0)* & 0.05 (0.0)* & -0.01 (0.36)* & 0.05 (0.0) \\
        32 & 0.02 (0.0)* & 0.05 (0.0)* & -0.02 (0.06)* & 0.04 (0.0)* \\
        33 & 0.01 (0.31)* & 0.06 (0.0)* & 0.0 (0.9)* & 0.06 (0.0) \\     \bottomrule  
        \end{tabular}
    \caption{Layer-wise RSA values (Spearman correlations) for the sentence condition and visual network, with $p$-values indicating the significance of the correlation (i.e., whether it is different from 0) in brackets. Note that asterisks indicate whether each correlation is statistically significantly different from the correlation achieved by the best layer (boldfaced value).}
    \label{tab:visual_large_models}
\end{table*}

\begin{table*}
    \centering\
    \scalefont{0.75}
    \begin{tabular}{llllllll} \toprule
        Layer & LXMERT & VisualBERT & BERT & LLaVA & IDEFICS2 & Llama3 & Mistral \\ \midrule
        1 & 0.23 (0.0)* & 0.33 (0.0) & -0.01 (0.11)* & 0.04 (0.0)* & -0.05 (0.0)* & -0.11 (0.0)* & -0.12 (0.0)*  \\
        2 & 0.23 (0.0)* & 0.31 (0.0) & \textbf{0.11} (0.0) & 0.05 (0.0)* & 0.06 (0.0)* & -0.1 (0.0)* & -0.1 (0.0)*  \\
        3 & 0.24 (0.0)* & 0.32 (0.0) & 0.06 (0.0)* & -0.08 (0.0)* & -0.0 (0.56)* & -0.13 (0.0)* & -0.13 (0.0)*  \\
        4 & 0.23 (0.0)* & 0.32 (0.0) & 0.07 (0.0)* & -0.0 (0.83)* & -0.04 (0.0)* & -0.11 (0.0)* & -0.1 (0.0)*  \\
        5 & 0.23 (0.0)* & 0.32 (0.0) & 0.01 (0.5)* & 0.03 (0.0)* & -0.04 (0.0)* & -0.15 (0.0)* & -0.09 (0.0)*  \\
        6 & 0.26 (0.0)* & 0.31 (0.0) & -0.07 (0.0)* & 0.07 (0.0)* & -0.03 (0.0)* & -0.13 (0.0)* & -0.05 (0.0)*  \\
        7 & 0.26 (0.0)* & 0.3 (0.0)* & -0.08 (0.0)* & 0.1 (0.0)* & -0.06 (0.0)* & -0.13 (0.0)* & -0.06 (0.0)*  \\
        8 & 0.28 (0.0)* & 0.31 (0.0) & -0.07 (0.0)* & 0.09 (0.0)* & -0.04 (0.0)* & -0.13 (0.0)* & -0.04 (0.0)*  \\
        9 & 0.27 (0.0)* & 0.31 (0.0)* & -0.04 (0.0)* & 0.08 (0.0)* & -0.06 (0.0)* & -0.13 (0.0)* & -0.05 (0.0)*  \\
        10 & 0.27 (0.0)* & 0.31 (0.0)* & -0.03 (0.0)* & 0.11 (0.0)* & -0.01 (0.08)* & -0.13 (0.0)* & -0.05 (0.0)*  \\
        11 & 0.27 (0.0)* & 0.32 (0.0) & -0.02 (0.01)* & 0.09 (0.0)* & -0.01 (0.06)* & -0.13 (0.0)* & -0.07 (0.0)*  \\
        12 & 0.28 (0.0)* & \textbf{0.33} (0.0) & 0.02 (0.05)* & 0.07 (0.0)* & -0.03 (0.0)* & -0.13 (0.0)* & -0.08 (0.0)*  \\
        13 & 0.27 (0.0)* & 0.33 (0.0) & 0.01 (0.16)* & 0.04 (0.0)* & -0.04 (0.0)* & -0.13 (0.0)* & -0.09 (0.0)*  \\
        14 & \textbf{0.33} (0.0) & - & - & 0.07 (0.0)* & -0.04 (0.0)* & -0.13 (0.0)* & -0.1 (0.0)*  \\
        15 & - & - & - & 0.07 (0.0)* & -0.04 (0.0)* & -0.13 (0.0)* & -0.1 (0.0)*  \\
        16 & - & - & - & -0.02 (0.03)* & -0.1 (0.0)* & -0.13 (0.0)* & -0.14 (0.0)*  \\
        17 & - & - & - & -0.02 (0.06)* & -0.08 (0.0)* & -0.14 (0.0)* & -0.16 (0.0)*  \\
        18 & - & - & - & 0.02 (0.04)* & -0.14 (0.0)* & -0.14 (0.0)* & -0.19 (0.0)*  \\
        19 & - & - & - & -0.02 (0.01)* & -0.1 (0.0)* & -0.13 (0.0)* & -0.17 (0.0)*  \\
        20 & - & - & - & -0.03 (0.0)* & -0.06 (0.0)* & -0.12 (0.0)* & -0.15 (0.0)*  \\
        21 & - & - & - & -0.05 (0.0)* & -0.01 (0.08)* & -0.11 (0.0)* & -0.16 (0.0)*  \\
        22 & - & - & - & -0.04 (0.0)* & 0.04 (0.0)* & -0.09 (0.0)* & -0.15 (0.0)*  \\
        23 & - & - & - & 0.03 (0.0)* & 0.12 (0.0)* & -0.07 (0.0) & -0.11 (0.0)*  \\
        24 & - & - & - & 0.06 (0.0)* & 0.14 (0.0) & -0.07 (0.0) & -0.09 (0.0)*  \\
        25 & - & - & - & 0.09 (0.0)* & 0.15 (0.0) & -0.07 (0.0) & -0.06 (0.0)*  \\
        26 & - & - & - & 0.1 (0.0)* & 0.16 (0.0) & -0.07 (0.0) & -0.04 (0.0)*  \\
        27 & - & - & - & 0.09 (0.0)* & \textbf{0.16} (0.0) & -0.06 (0.0) & -0.02 (0.0)*  \\
        28 & - & - & - & 0.09 (0.0)* & 0.15 (0.0) & -0.07 (0.0) & -0.01 (0.07)*  \\
        29 & - & - & - & 0.1 (0.0)* & 0.14 (0.0)* & -0.06 (0.0) & -0.01 (0.07)*  \\
        30 & - & - & - & 0.12 (0.0)* & 0.14 (0.0)* & -0.07 (0.0) & -0.04 (0.0)*  \\
        31 & - & - & - & \textbf{0.18} (0.0) & 0.14 (0.0) & -0.07 (0.0) & -0.04 (0.0)*  \\
        32 & - & - & - & 0.17 (0.0) & 0.14 (0.0)* & -0.08 (0.0)* & -0.06 (0.0)*  \\
        33 & - & - & - & 0.1 (0.0)* & 0.1 (0.0)* & \textbf{-0.05} (0.0) & \textbf{0.04} (0.0)  \\
     \bottomrule  
        \end{tabular}
    \caption{Layer-wise RSA values (Spearman correlations) for the picture condition and LH language network, with $p$-values indicating the significance of the correlation (i.e., whether it is different from 0) in brackets. Note that asterisks indicate whether each correlation is statistically significantly different from the correlation achieved by the best layer (boldfaced value).}
    \label{tab:language_LH_large_models_picture}
\end{table*}

\begin{table*}
    \centering\
    \scalefont{0.75}
    \begin{tabular}{llllllll} \toprule
        Layer & LXMERT & VisualBERT & BERT & LLaVA & IDEFICS2 & Llama3 & Mistral \\ \midrule
        1 & 0.2 (0.0)* & \textbf{0.48} (0.0) & -0.01 (0.36)* & -0.02 (0.02)* & 0.01 (0.08)* & -0.07 (0.0)* & -0.04 (0.0)*  \\
        2 & 0.2 (0.0)* & 0.47 (0.0) & \textbf{0.08} (0.0) & 0.04 (0.0)* & 0.1 (0.0)* & -0.03 (0.0)* & -0.02 (0.0)*  \\
        3 & 0.21 (0.0)* & 0.46 (0.0)* & 0.06 (0.0) & -0.0 (0.72)* & 0.05 (0.0)* & -0.04 (0.0)* & -0.09 (0.0)*  \\
        4 & 0.2 (0.0)* & 0.46 (0.0)* & 0.06 (0.0) & 0.04 (0.0)* & 0.05 (0.0)* & -0.03 (0.0)* & -0.05 (0.0)*  \\
        5 & 0.19 (0.0)* & 0.45 (0.0)* & 0.02 (0.02)* & 0.06 (0.0)* & 0.04 (0.0)* & -0.05 (0.0)* & -0.04 (0.0)*  \\
        6 & 0.21 (0.0)* & 0.43 (0.0)* & -0.03 (0.0)* & 0.08 (0.0)* & 0.05 (0.0)* & -0.05 (0.0)* & 0.0 (0.72)*  \\
        7 & 0.22 (0.0)* & 0.42 (0.0)* & -0.04 (0.0)* & 0.11 (0.0)* & 0.03 (0.0)* & -0.05 (0.0)* & -0.02 (0.03)*  \\
        8 & 0.22 (0.0)* & 0.43 (0.0)* & -0.02 (0.01)* & 0.11 (0.0)* & 0.02 (0.04)* & -0.05 (0.0)* & 0.01 (0.35)*  \\
        9 & 0.22 (0.0)* & 0.42 (0.0)* & 0.02 (0.03)* & 0.11 (0.0)* & 0.01 (0.07)* & -0.05 (0.0)* & -0.0 (0.62)*  \\
        10 & 0.22 (0.0)* & 0.41 (0.0)* & -0.0 (0.91)* & 0.14 (0.0)* & 0.03 (0.0)* & -0.05 (0.0)* & -0.01 (0.46)*  \\
        11 & 0.25 (0.0)* & 0.4 (0.0)* & -0.01 (0.48)* & 0.13 (0.0)* & 0.03 (0.0)* & -0.05 (0.0)* & -0.02 (0.01)*  \\
        12 & 0.27 (0.0) & 0.39 (0.0)* & 0.02 (0.0)* & 0.12 (0.0)* & 0.02 (0.01)* & -0.05 (0.0)* & -0.02 (0.0)*  \\
        13 & 0.27 (0.0) & 0.36 (0.0)* & 0.03 (0.0)* & 0.08 (0.0)* & 0.03 (0.0)* & -0.05 (0.0)* & -0.03 (0.0)*  \\
        14 & \textbf{0.29} (0.0) & - & - & 0.1 (0.0)* & 0.03 (0.0)* & -0.05 (0.0)* & -0.04 (0.0)*  \\
        15 & - & - & - & 0.08 (0.0)* & 0.02 (0.0)* & -0.05 (0.0)* & -0.04 (0.0)*  \\
        16 & - & - & - & 0.04 (0.0)* & -0.01 (0.43)* & -0.05 (0.0)* & -0.06 (0.0)*  \\
        17 & - & - & - & 0.05 (0.0)* & 0.0 (0.84)* & -0.05 (0.0)* & -0.06 (0.0)*  \\
        18 & - & - & - & 0.04 (0.0)* & -0.03 (0.0)* & -0.05 (0.0)* & -0.09 (0.0)*  \\
        19 & - & - & - & 0.05 (0.0)* & -0.01 (0.12)* & -0.04 (0.0)* & -0.08 (0.0)*  \\
        20 & - & - & - & 0.02 (0.01)* & 0.0 (0.96)* & -0.05 (0.0)* & -0.08 (0.0)*  \\
        21 & - & - & - & 0.02 (0.01)* & 0.07 (0.0)* & -0.05 (0.0)* & -0.05 (0.0)*  \\
        22 & - & - & - & 0.04 (0.0)* & 0.08 (0.0)* & -0.02 (0.01)* & -0.08 (0.0)*  \\
        23 & - & - & - & 0.07 (0.0)* & 0.11 (0.0) & -0.0 (0.6)* & -0.06 (0.0)*  \\
        24 & - & - & - & 0.09 (0.0)* & 0.12 (0.0) & -0.0 (0.98)* & -0.05 (0.0)*  \\
        25 & - & - & - & 0.1 (0.0)* & 0.13 (0.0) & 0.0 (0.63)* & -0.03 (0.0)*  \\
        26 & - & - & - & 0.1 (0.0)* & 0.13 (0.0) & 0.0 (0.57)* & -0.02 (0.03)*  \\
        27 & - & - & - & 0.11 (0.0)* & \textbf{0.13} (0.0) & 0.01 (0.29)* & -0.0 (0.67)*  \\
        28 & - & - & - & 0.11 (0.0)* & 0.11 (0.0) & 0.01 (0.41)* & 0.01 (0.4)*  \\
        29 & - & - & - & 0.12 (0.0)* & 0.1 (0.0)* & 0.01 (0.41)* & 0.01 (0.19)*  \\
        30 & - & - & - & 0.13 (0.0)* & 0.09 (0.0)* & 0.0 (0.59)* & 0.0 (0.97)*  \\
        31 & - & - & - & \textbf{0.17} (0.0) & 0.11 (0.0)* & 0.0 (0.71)* & -0.01 (0.4)*  \\
        32 & - & - & - & 0.17 (0.0) & 0.1 (0.0)* & -0.0 (0.67)* & -0.02 (0.05)*  \\
        33 & - & - & - & 0.17 (0.0) & 0.08 (0.0)* & \textbf{0.08} (0.0) & \textbf{0.05} (0.0)  \\
         \bottomrule  
        \end{tabular}
    \caption{Layer-wise RSA values (Spearman correlations) for the picture condition and visual network, with $p$-values indicating the significance of the correlation (i.e., whether it is different from 0) in brackets. Note that asterisks indicate whether each correlation is statistically significantly different from the correlation achieved by the best layer (boldfaced value).}
    \label{tab:visual_large_models_picture}
\end{table*}

\subsection{Picture condition ablation: Additional comparison}
\label{sec:appendix_pic_cond_ablation}

\begin{figure*}
    \centering
    \includegraphics[width=0.8\linewidth]{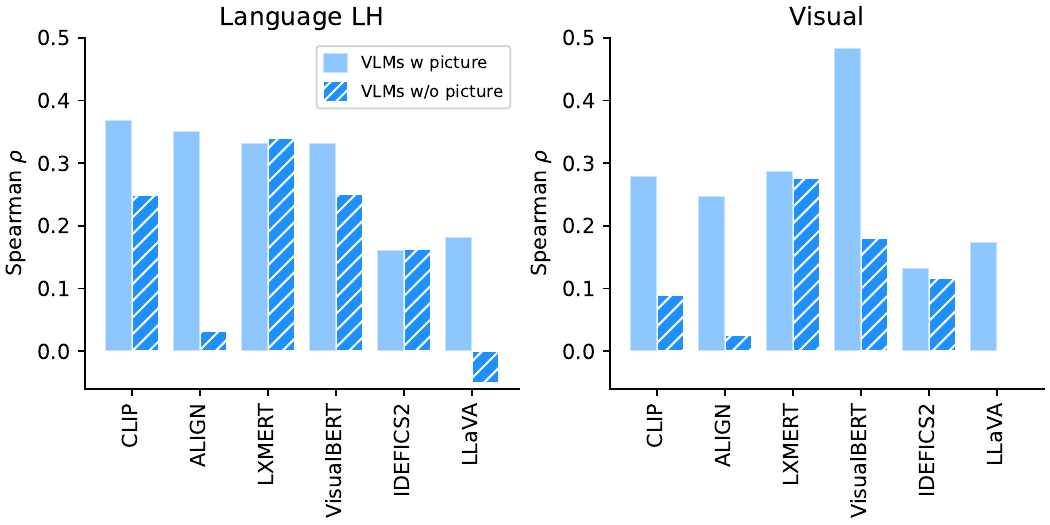}
    \caption{Initial VLM results from RSA analysis in the picture condition vs.~results obtained when passing only concept words (without images).}
    \label{fig:pic-cond-ablation}
\end{figure*}

To facilitate comparison between VLMs' brain alignment in the picture condition when the image is passed vs.~when it is not, we visualise this information in Figure~\ref{fig:pic-cond-ablation}.

\end{document}